\documentclass[journal]{IEEEtran}
\ifCLASSOPTIONcompsoc

  \usepackage[nocompress]{cite}
\else
  \usepackage{cite}
\fi

\usepackage{bbm}
	
\usepackage[T1]{fontenc}

\ifCLASSINFOpdf

\else

\fi

\usepackage{hyperref}
\usepackage{graphicx}
\usepackage{array}
\newcolumntype{P}[1]{>{\centering\arraybackslash}p{#1}}
\newcolumntype{M}[1]{>{\centering\arraybackslash}m{#1}}
\begin{document}
%
\title{Predicting visual attention in graphic design documents}
%
%
%
%

\author{Souradeep Chakraborty,
        Zijun Wei, Conor Kelton, Seoyoung Ahn, Aruna Balasubramanian, Gregory J. Zelinsky,
        and~Dimitris Samaras
\thanks{S. Chakraborty (Email: souchakrabor@cs.stonybrook.edu), Z. Wei, C. Kelton, A. Balasubramanian, G.J. Zelinsky and D. Samaras are with Department of Computer Science, Stony Brook University, USA.}
\thanks{G.J. Zelinsky and S. Young are with Department of Psychology, Stony Brook University, USA.}
}

%
%

\markboth{Journal of \LaTeX\ Class Files,~Vol.~14, No.~8, August~2015}%
{Shell \MakeLowercase{\textit{et al.}}: Bare Demo of IEEEtran.cls for Computer Society Journals}
%




\IEEEtitleabstractindextext{%
\begin{abstract}
We present a model for predicting visual attention during the free viewing of graphic design documents. While existing works on this topic have aimed at predicting static saliency of graphic designs, our work is the first attempt to predict both spatial attention and dynamic temporal order in which the document regions are fixated by gaze using a deep learning based model. We propose a two-stage model for predicting dynamic attention on such documents, with webpages being our primary choice of document design for demonstration. In the first stage, we predict  the saliency maps for each of the document components (e.g. logos, banners, texts, etc. for webpages) conditioned on the type of document layout. These component saliency maps are then jointly used to predict the overall document saliency. In the second stage, we use these layout-specific component saliency maps as the state representation for an inverse reinforcement learning model of fixation scanpath prediction during document viewing. To test our model, we collected a new dataset consisting of eye movements from 41 people freely viewing 450 webpages (the largest dataset of its kind). Experimental results show that our model outperforms existing models in both saliency and scanpath prediction for webpages, and also generalizes very well to other graphic design documents such as comics, posters, mobile UIs, etc. and natural images.
\end{abstract}

}

\maketitle

\IEEEdisplaynontitleabstractindextext

%
\IEEEpeerreviewmaketitle


\section{Introduction}

Unlike natural images, graphic design documents are created by people to convey information, such as flyers, webpages, etc. This means that their layout  usually  directs a viewer's attention to particular elements of the design, perhaps in a specific viewing order. Predicting a user's attention is therefore critically important to knowing whether the graphic design  efficiently conveys the intended information~\cite{bylinskii2017learning}. Also unlike natural images, graphic designs try to direct attention using a mixture of captivating text fonts, animations, and \textit{images}. Competition between these design elements for attention makes this a more difficult problem than attention prediction for natural images, not only because images are included in documents, but also because complex cognitive processes are engaged (e.g., reading) that are not part of most natural image viewing. 

Previous methods to predict attention allocation  during document viewing~\cite{shen2014webpage,li2016webpage,gu2019element} have significant limitations. For one, they mainly adopt black-box solutions that lack interpretability, making it difficult to understand how individual document components contribute to attention behavior. Also, these works have largely attempted to predict only the spatial distribution of attention over a document, as estimated by purely spatial fixation-density maps (FDMs). However, visual attention is a dynamic process that changes both spatially as well as over time. The regions of a document can be attended in different orders~\cite{xia2020predicting} which likely signify different levels of engagement with the document.  

Predicting the dynamic allocation of a person's attention can have far-reaching applications. For one, growth in remote education has created the need for predictive models of attention in the context of online meeting platforms, which are also graphic documents. If one could predict attention shift between speakers, shared image content (i.e., slides), etc., then intelligent systems~\cite{lagun2016understanding} could anticipate a viewer's attention and introduce content to maximize a desired outcome (e.g., comprehension of presented material). At a shorter timescale, being able to predict attention, even for the first few gaze fixations, could enable preferential content loading, thereby making delays resulting from bandwidth limitations less disruptive~\cite{webgaze2017}. Critically, however, existing  saliency prediction methods \textit{do not} consider this temporal dynamic. 

We fill this gap by introducing \textit{Attention on Graphic Designs (AGD)}, a model for predicting the  spatio-temporal allocation of visual attention (gaze scanpaths) during graphic design document viewing. Our primary focus is on \textit{webpages}, due to their diversity of components (texts, logos, images etc.) making these documents a challenging test of saliency models~\cite{faraday2000visually,shen2014webpage}, but we show that our method generalizes beyond webpages to other design documents such as comics, posters, and mobile user interfaces (Sec.~\ref{sec:generalization}).

\textit{AGD} consists of two highly-interpretable stages. The first stage, AGD-Fixation (AGD-F), is related to existing work in that we use a saliency map to predict the FDM (the aggregate spatial distribution of a viewer's fixations over the document) and we compute saliency using face, text, and other document image features. However, whereas previous models assumed that all text and faces are in a sense equally salient, our model uses only the faces and text that \textit{attract the most attention}. Prioritizing the importance of individual text and faces is a contribution of our work because documents often depict multiple instances of both, and not all instances attract attention equally well (see Sec.~\ref{sec:salcomp}). Our approach thus  models attention as a combination of only the \textit{salient} instances within multiple document components. 

In addition to salience-weighted combination of document components, another innovation of our study is the use of \textit{page layout} as another feature for predicting gaze, adding to the features that are specific to the different document components. The design of a document, and specifically its layout, is likely to impact a viewer's  attention allocation, which can be biased even by (content-less) geometric shapes \cite{findlay1982global}. Existing saliency methods \cite{shen2014webpage,fosco2020predicting} do not consider page layout in their predictive models, assuming that a layout bias would simply disappear when these boxes are filled with content, as in a webpage. Here we assume otherwise, that different design layouts can lead to  different document viewing behaviors. We include layout information in our model by passing a layout embedding directly as an input to the model encoder, thus factoring webpage layout into the overall saliency prediction. We obtained webpage layouts using page metadata and purely data-driven segmentation methods (Sec.~\ref{sec:stage1}) over a large repository of 55,000 popular webpages. 

Webpage FDMs are computed in stage 1 so that they can be used as the state representation in the stage 2 modeling aimed at predicting the temporal order of fixations over different webpage regions, referred to here as webpage scanpaths. For a modeling approach we use inverse reinforcement learning (IRL), which was recently shown to predict scanpaths in the context of search tasks and natural images~\cite{yang2020predicting}. We extend this model to the free viewing of webpages by adopting a state representation consisting of a webpage's semantic elements (texts, banners, etc.). We call this model AGD-Scanpath (AGD-S). By integrating a component-specific fixation-prediction network with a modified inverse-reinforcement learning paradigm developed for natural images, we found that our model showed unprecedented generalization between different types of graphic designs. To the best of our knowledge, ours is the first deep learning model to predict fixation scanpaths on multiple graphic design types (e.g., webpages and mobile UIs).

One potential reason for the relatively poor performance shown by previous webpage attention prediction models ~\cite{shen2014webpage,li2016webpage,xia2020predicting} may be because these models were trained on datasets of insufficient size (the previously best dataset, FiWI, has 149 images). To address this limitation we created \textit{WebSaliency}, a gaze fixation dataset consisting of 450 webpage images sampled from our stage 1 layout clusters (to capture visual layout diversity). This makes WebSaliency the largest currently available dataset of free-viewing gaze fixations during webpage viewing, and another contribution. We show that WebSaliency exhibits more entropy variation than FiWI, and  models perform better if trained on WebSaliency. Related to WebSaliency is the IMP1K dataset \cite{fosco2020predicting}, which is a graphic
design dataset covering five design classes: infographics, webpages, mobile UIs, advertisements and movie posters. Similar to the data in the IMP1K dataset for webpage designs, the WebSaliency dataset we present in this paper is used to train models that can predict free-viewing attention prediction on webpages. The attention data in IMP1K comes in the form of importance maps that are  obtained using importance region annotations from Amazon Turk participants. Although the attention data in IMP1K provides us the relative importance of the semantic regions in a graphic design e.g. logos, images, banners, etc., similar to WebSaliency, it fails to provide finer details about how attention prioritization differs within a particular  semantic region, which is critical for a more accurate attention prediction. Our study shows that attention allocation can often have high spatial variance within a given semantic region, e.g. a block of text in a news website, which the IMP1K data often fails to capture due to the absence of more accurate eye fixation data (see Sec.  \ref{sec:experiments} for these results).\\

Perhaps the greatest contribution of our work, however, is in our use of a single model to predict two different types of attention behavior -- fixation-density maps and scanpaths. Previous models of attention prediction have been applied to the task of \textit{either} fixation-density map prediction or the task of predicting fixation scanpaths, but no single model exists that predicts both attention behaviors. Moreover, we found that our model outperforms the more specialized models built to predict one behavior or the other. In multiple experiments we demonstrate that the reason for our model's improved attention prediction performance stems from its explicit prediction of component-specific FDMs before predicting the final FDM improves saliency prediction performance. Existing state-of-the-art models, such as SAM-ResNet, DI Net, EML-NET etc., do not model component-specific saliency as an intermediate step in attention prediction.

\noindent To summarize these contributions:
\begin{itemize}
  \item The first unified and interpretable deep learning model for predicting static and dynamic visual attention behavior (fixation density maps and scanpaths) during the free viewing of different graphics design documents.   
  
  \item Demonstrations that saliency within document components and document layout information can be leveraged to improve attention prediction in graphic designs.
  
  \item Introducing the largest dataset (to date) of free-viewing gaze fixations on webpage images.
  
\end{itemize}

\begin{figure*}
\centering
\includegraphics[scale=0.22]{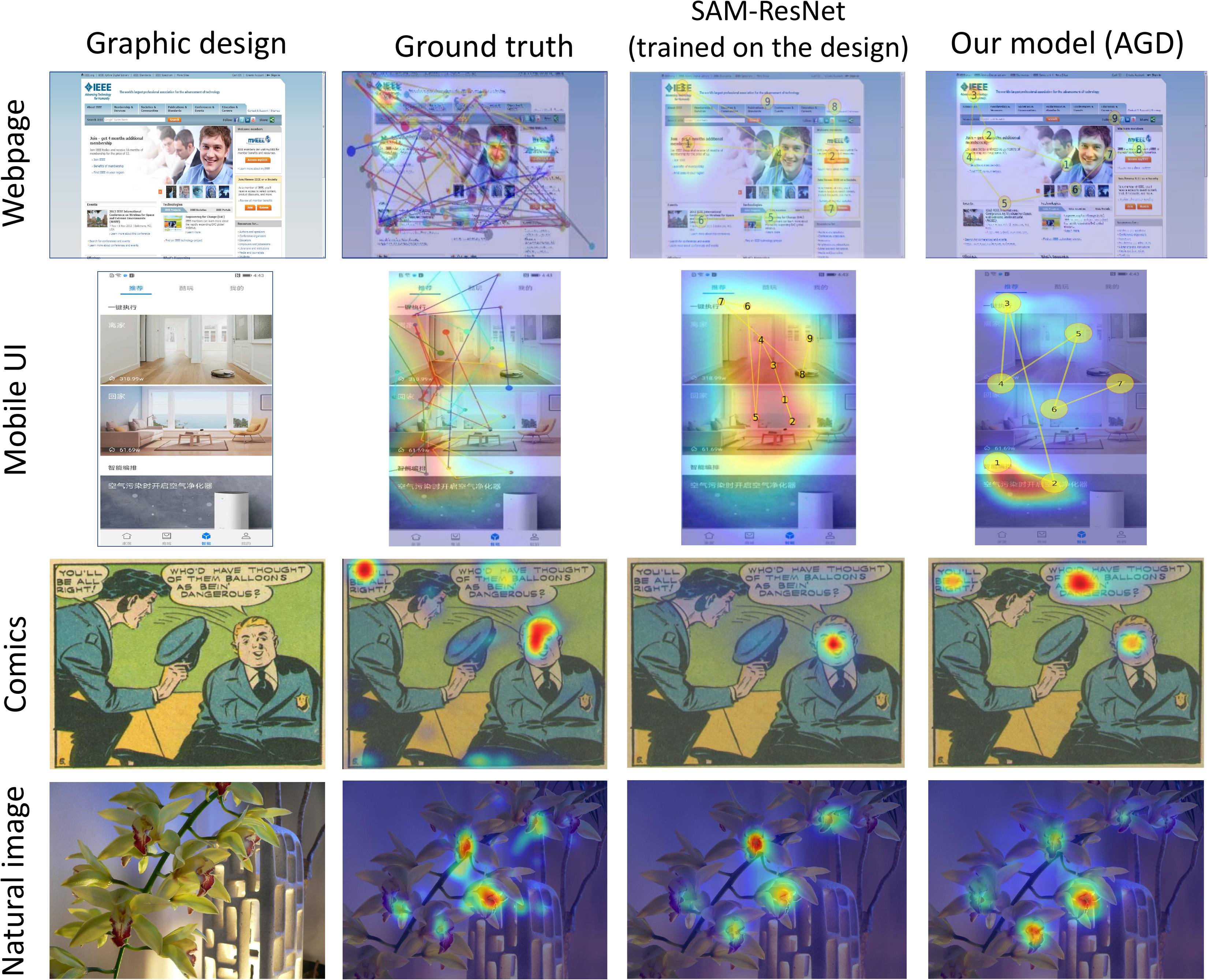}
\caption {Overlaid predicted fixation density maps of the SAM-ResNet model and our AGD model (trained on our WebSaliency dataset) on the \textit{ieee.org} webpage (from FiWI \cite{shen2014webpage}), a mobile UI image (from  Mobile UI \cite{gupta2018saliency}), a comics image (from DeepComics \cite{bannier2018deepcomics}) and a natural image (from MIT 1003 \cite{judd2009learning}) and scanpaths for the webpage and the mobile UI instance. Our predictions more closely resembles the ground truth and our model shows great generalization across different types of graphic designs although the model is trained to predict fixations on webpage images.}
\label{fig:teaser}
\end{figure*}

To foreshadow our results, Fig. \ref{fig:teaser} shows saliency predictions for different methods for four different types of graphic designs (webpage, mobile UI, comics and natural image) and scanpath predictions for the webpage and the mobile UI instance. Our model predicts fixations better compared to the natural saliency model SAM-ResNet \cite{cornia2018predicting}, even when SAM-ResNet is trained on these documents instead of natural images. Our AGD model demonstrates unprecedented generalization across different types of graphic designs even when the model is trained to predict fixations on webpage images only (using our WebSaliency dataset).

\vspace{-2.5mm}
\section{Related Work}
\label{sec:relatedwork}
\subsection{Attention Prediction on Natural Images}
Computer vision has made rapid progress in image saliency detection, especially using deep learning methods, which can be broadly classified into: bottom-up models \cite{itti1998model,harel2007graph,zhang2008sun,hou2007saliency,zhang2013saliency}
and top-down models \cite{yang2016top,kanan2009sun,kocak2014top,ramanishka2017top}. Notable works include the use of multi-scale deep features \cite{li2015visual}, multi-contextual features \cite{zhao2015saliency}, and  recurrent deep models  \cite{wang2016saliency,cornia2018predicting}. See  \cite{borji2019saliency} for a good review of saliency models and datasets in the deep learning era. Also, several works have aimed to predict scanpaths on natural images \cite{assens2018pathgan,xia2019predicting,sun2019visual}.  
These natural image attention models do not generalize well to graphic design documents, as shown in previous studies \cite{shen2014webpage,bylinskii2017learning}. This motivates us to build a model for such documents.

\begin{figure*}[h]
\centering
\includegraphics[width=18.2cm]{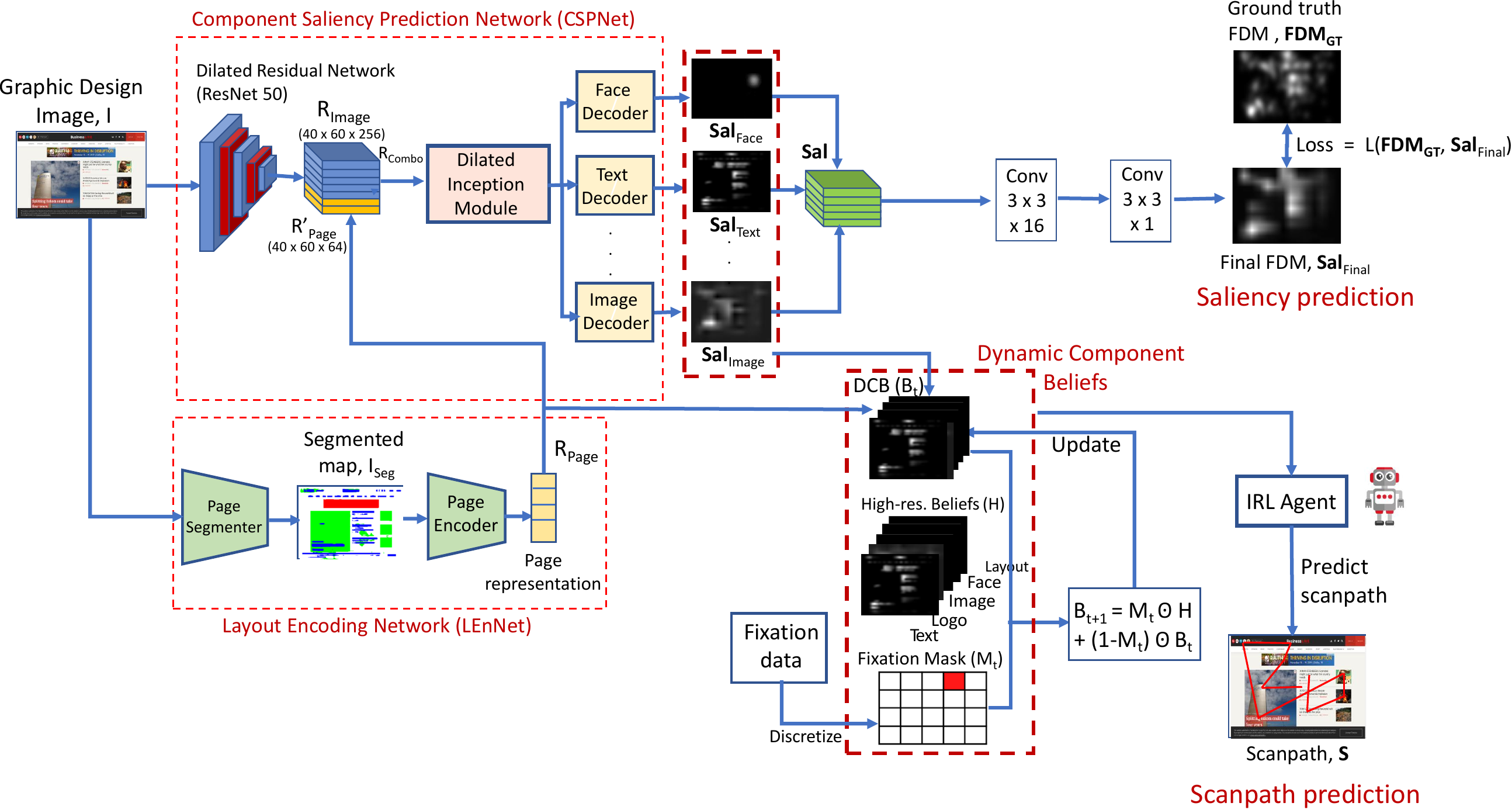}

\caption {Our graphic design attention prediction model, \textit{AGD}. For saliency map prediction (Stage 1), we first extract the image representations using a dilated residual network (with ResNet 50 as the backbone) and combine them with the page layout representations obtained using the Layout Encoding Network (LEnNet) to form the $R_{combo}$ representation,  which is input to the dilated inception module network as the encoder. Fixation density maps of the document components are then predicted using the corresponding image decoders and are combined and passed through two $3\times 3$ convolution layers to form the final saliency map, $Sal_{Final}$. For scanpath prediction (Stage 2), the component FDMs from Stage 1 are used for constructing the dynamic component belief maps, which are discretized sequentially into binary fixation masks. The state of  the inverse reinforcement learning (IRL) model is updated as depicted. The IRL agent is trained to predict the scanpath, S by applying  Inhibition-of-Return on the intermediate fixation probability map (see Section  \ref{sec:scanpath_method} for details).}
\label{fig:flowchart2}
\end{figure*}

\vspace{-2.5mm}
\subsection{Attention Prediction on Graphic Designs}
There have been several works that have predicted saliency on graphic designs such as webpages, posters, etc. Shen et al. \cite{shen2014webpage,shen2015predicting} combined hand-crafted features (e.g. text, positional bias, etc.) with straightforward fitting of these feature maps to predict eye fixations on webpages. In comparison, our model first finds the \textit{saliency} of the visual components and then combines these salient components. Zheng et al.~\cite{zheng2018task} collected human fixations under various tasks such as shopping and form-filling, and predicted saliency based on these task labels.
Gu et al.~\cite{gu2019element} captured page biases using a variational autoencoder and used feature detectors to predict webpage saliency. Xia et al.'s \ \cite{xia2020predicting} saccadic model predicts scanpaths while viewing webpages, using  hand-crafted features for training and is not interpretable. We,  however, adopt a deep learning approach that better reveals the dynamic attention mechanisms,  being trained on a larger dataset of gaze fixations. This helps us improve prediction performance with interpretable results.

Bylinskii et al.~\cite{bylinskii2017learning} used deep learning  to predict saliency for data visualizations and graphics. Deka et al.~\cite{deka2017rico} collected a dataset  of mobile UIs and learned a low dimensional representation using an AutoEncoder. In a similar spirit, we have  encoded document layouts. Further along \cite{gupta2018saliency,leiva2020understanding} investigated saliency in mobile UIs. Bannier et al. \cite{bannier2018deepcomics} applied deep networks in saliency prediction on  comics. Recently, \cite{fosco2020predicting} presented a model that first classifies a graphic design and then predicts design-specific importance. However, this work predicts importance maps, collected using mouse annotations instead of eye fixation data. While we compare our model with the model in \cite{fosco2020predicting}, our findings suggest that the importance maps obtained by their model do not well predict the eye fixations in our WebSaliency dataset due to qualitative differences. The model we 
propose generalizes to saliency prediction for posters, comics, mobile UIs and natural images.
\vspace{-1.8mm}
\section{Proposed Method}
\label{sec:method}
\subsection{Model Overview}
We  propose  a  two-stage  model for predicting attention on graphic design documents. In the first stage, we predict the fixation density map (FDM) by first computing the FDMs of the components that constitute a graphic design such as faces, texts, banners, etc., using a dilated inception module as the encoder that leverages image features and an explicit page layout representation.  Next, these component FDMs are used to predict the final FDM. In the second stage, we leverage the component FDMs computed in Stage 1 to predict the visual scanpath (sequence of fixations on the document image) using an inverse reinforcement learning (IRL) approach. We describe in details these stages in the following subsections.
\vspace{-2mm}
\subsection{Stage 1: Fixation Density Map Prediction}
\label{sec:stage1}
\noindent As is well known in graphic designs, human related features, like eyes and faces~\cite{cerf2008predicting,judd2009learning}, logos, and even text~\cite{shen2014webpage} attract more attention than other objects. Thus, we design our model to include information from all of these sources, and to find the overall importance of these features relative to one another. Our results also show that not every face or textual region is important in attracting attention. We thus opt to use face and text \textit{saliency} maps instead of their raw feature encodings to predict attention. Similarly, we use logo and banner saliency maps in our prediction model as these components have also been found to influence attention in designs such as webpages~\cite{shen2014webpage}. Our model also includes a natural image saliency map that captures attention for other visually discriminative regions. 

Fig.~\ref{fig:flowchart2} shows AGD, our saliency and scanpath prediction model for graphic design documents. In stage 1 (AGD-F), the predicted component FDMs capture the saliency of faces, texts, logos, banners and other image regions. We use the feature combination  
[$Sal_{Face},Sal_{Text},Sal_{Logo},Sal_{Banner},Sal_{Img}$], conditioned on the page layout   $L_{Page}$, that best approximates the ground truth data, $FDM_{GT}$ using $Sal_{Final}$, the final saliency prediction map of our AGD-F method. In stage 2  (AGD-S), the input image and its low-resolution counterpart are converted into the corresponding component belief maps which are leveraged to predict scanpaths using generative adversarial imitation learning (detailed in Sec.~\ref{sec:scanpath_method}).

\vspace {0.8mm}
\noindent \textbf{Component Saliency Prediction:}
\label{sec:salcomp}
We propose a Component Saliency Prediction Network (CSPNet), a deep network for predicting the fixation density maps of the components that constitute a graphic design. More specifically, the face, text, logo, banner and image FDMs, constitute the final FDM. CSPNet predicts the saliency of each of these components.

Several works \cite{he2019understanding,kummerer2016deepgaze} have used convolutional networks with encoder-decoder architectures to predict saliency in natural images. \cite{yang2019dilated} has shown that using dilated convolutions in an encoder-decoder model  better captures the multi-scale contextual information in an image and improves saliency prediction. Motivated by the success of this model, we first encode the document image using a dilated residual network (with ResNet-50 backbone) \cite{liu2018deep,cornia2018predicting}, which we used as our primary feature extractor. We add a dilated inception module on top of the backbone network for capturing the multi-scale image features, following \cite{yang2019dilated}.  

Next, we predict the component FDMs using separate decoders dedicated to each document component. Each decoder unit consists of three stacked  convolutional layers with
a bilinear up-sampling layer at the end. The sub-networks for the prediction of the saliency of the components are jointly trained by combining the loss functions corresponding to the five components as:
    $L_{combo} = \lambda_{F} L_{F}+ \lambda_{T} L_{T}+ \lambda_{L} L_{L}+ \lambda_{B} L_{B}+ \lambda_{I} L_{I} $
where  $\lambda_{F (Face)}$ = $3$, $\lambda_{T (Text)}$ = $1$,$\lambda_{L (Logo)}$ = $2$,$\lambda_{B (Banner)}$ = $1$, $\lambda_{I (Image)}$ = $1$, are set empirically  based on our  validation set (see supplementary for details). 

\begin{figure}
\centering
\includegraphics[width = 8.3cm]{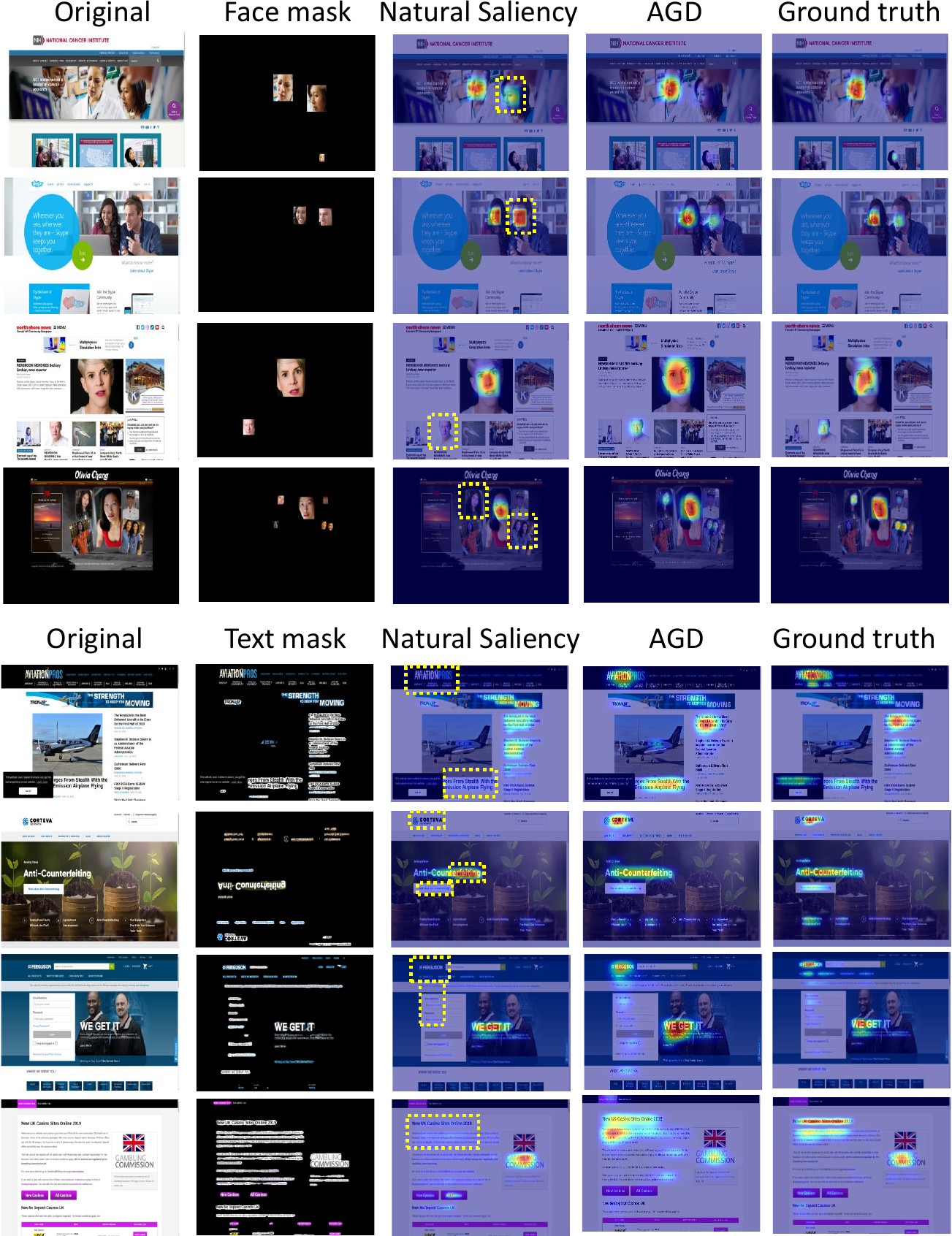}
\caption {Comparison of predicted face and text FDMs of a natural saliency model UAVDVSM \cite{he2019understanding} and AGD-F. Striped yellow boxes indicate the inaccuracies in UAVDVSM predictions. AGD-F produces more accurate salient component FDMs.} 
\label{fig:facetextmasks}
\vspace{-2.5mm}
\end{figure}

Finally, the FDMs predicted by the decoders are concatenated into a single volume and two $3\times3$ convolutional layers, C1 and C2 with 16 and 1 channels respectively are used to map the predicted component FDMs into the  final FDM. The input to the C1 layer is the volume constructed by concatenating the predicted component FDMs of the CSPNet sub-network. The output of the C2 layer is the final predicted FDM. 

The loss function we used for training our networks (for both predicted component FDMs and the final FDM) is composed of: (1) Total variation loss \cite{verdu2014total}, $L_{TV}$ and (2) Normalized Scanpath Saliency loss, $L_{NSS}$. The NSS loss was computed as $L_{NSS} = 1/S_{NSS}$, where $S_{NSS}$ is the NSS score of the predicted map. Specifically, total loss, $L_{Total} = \lambda_{TV}L_{TV} + \lambda_{NSS}L_{NSS}$, where $\lambda_{TV} = 0.7$, and $\lambda_{NSS} = 0.3$ are set based on our validation dataset (see supplementary for details).

To compute the salient text regions, first we detected textual regions using a text detector  ~\cite{long2018textsnake} and computed the text mask map, $Map_{Text}$. Given the text locations, we obtain the ground truth Text FDM, $FDM_{Text}$ by multiplying the text mask, $Map_{Text}$ and the ground truth FDM, $FDM_{GT}$ (from  WebSaliency) pixelwise. We use $FDM_{Text}$ as the target map for the text sub-network of CSPNet during training. At inference, the predicted text FDM, $FDM_{Text}$ is convolved with a Gaussian kernel, $G_{kern}$ with standard deviation, $\sigma = 25$ pixels to generate the final predicted text FDM, $Sal_{Text}$. 50 pixels correspond to 1 visual degree of foveal region in the human eye \cite{shen2014webpage}. 
To extract the salient faces we use the pre-trained publicly available CNN based \textit{Dlib} face detection method \cite{sharma2016farec}. The face FDM, $Sal_{Face}$ is generated  similarly to the text FDM. For capturing the saliency of logos, we manually annotated logo regions in our dataset (as they could not be directly obtained from the page metadata) and trained a UNet++ model \cite{zhou2018unet++} to segment logos in webpage images. The banner regions were obtained automatically using our PageSegNet segmentations. The logo and the banner FDMs were obtained similarly to the other component FDMs. We captured the image saliency due to other discriminative image regions by first predicting an overall image FDM, $Sal_{Image}^{whole}$ and then masking the other component regions in this map to obtain the residual FDM  $Sal_{Image}$, which captures the saliency due to discriminative image regions.

In Fig.~\ref{fig:facetextmasks}, we visually compare the face and text saliency prediction performance of our method with UAVDVSM~\cite{he2019understanding}, a saliency prediction model for natural images. In cases where multiple faces are present, UAVDVSM either predicts all faces to be equally salient (rows 1 and 2) or incorrectly allocates attention to certain faces (rows 3 and 4). Similarly for text, UAVDVSM fails to predict the text saliency in logos (rows 1 to 3) and is otherwise inaccurate for texts (rows 1, 2 and 4).

\vspace{1.2mm}
\noindent\textbf{Automatic webpage segmentation:} Visual segmentation of  graphical stimuli, such as webpages or mobile apps, is a well studied process in UI communities~\cite{pang2016flow,webgaze2017,deka2017rico}. We script the web browser to obtain automated access to the Document Object Models (DOM) and CSS Object Models (CSSOM) of the webpages. We use these models to segment the pages into their components, including HTML images, advertisements, and paragraphs. We selected 55K pages from the 1 million most popular webpages~\cite{jones2012majestic}. See supplementary for details regarding the  segmentation of different webpage components.

\begin{figure}
\centering
\includegraphics[width=8.8cm,height = 3.2cm]{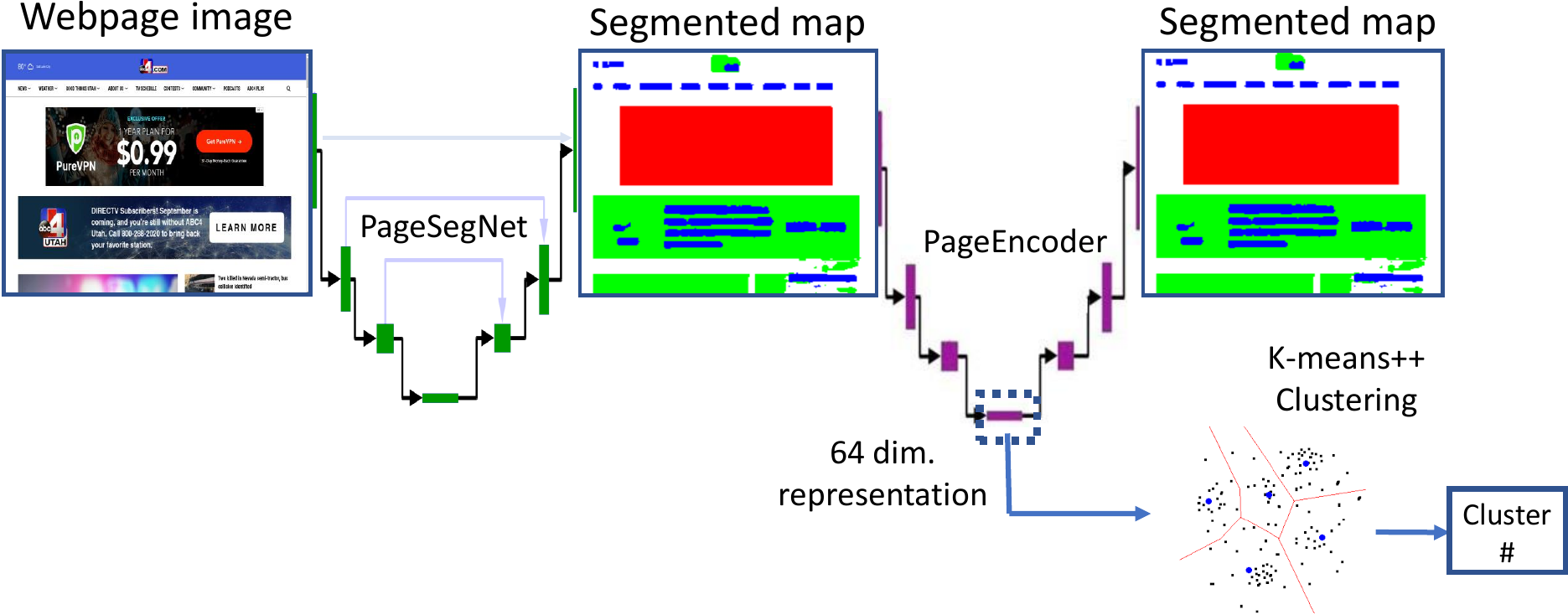}
\caption {Pipeline to obtain the page cluster from the original image using connected PageSegNet and PageEncoder networks.} 
\label{fig:segs}
\vspace{-1.5mm}
\end{figure}

\begin{figure}
\centering
\includegraphics[width=8.9cm]{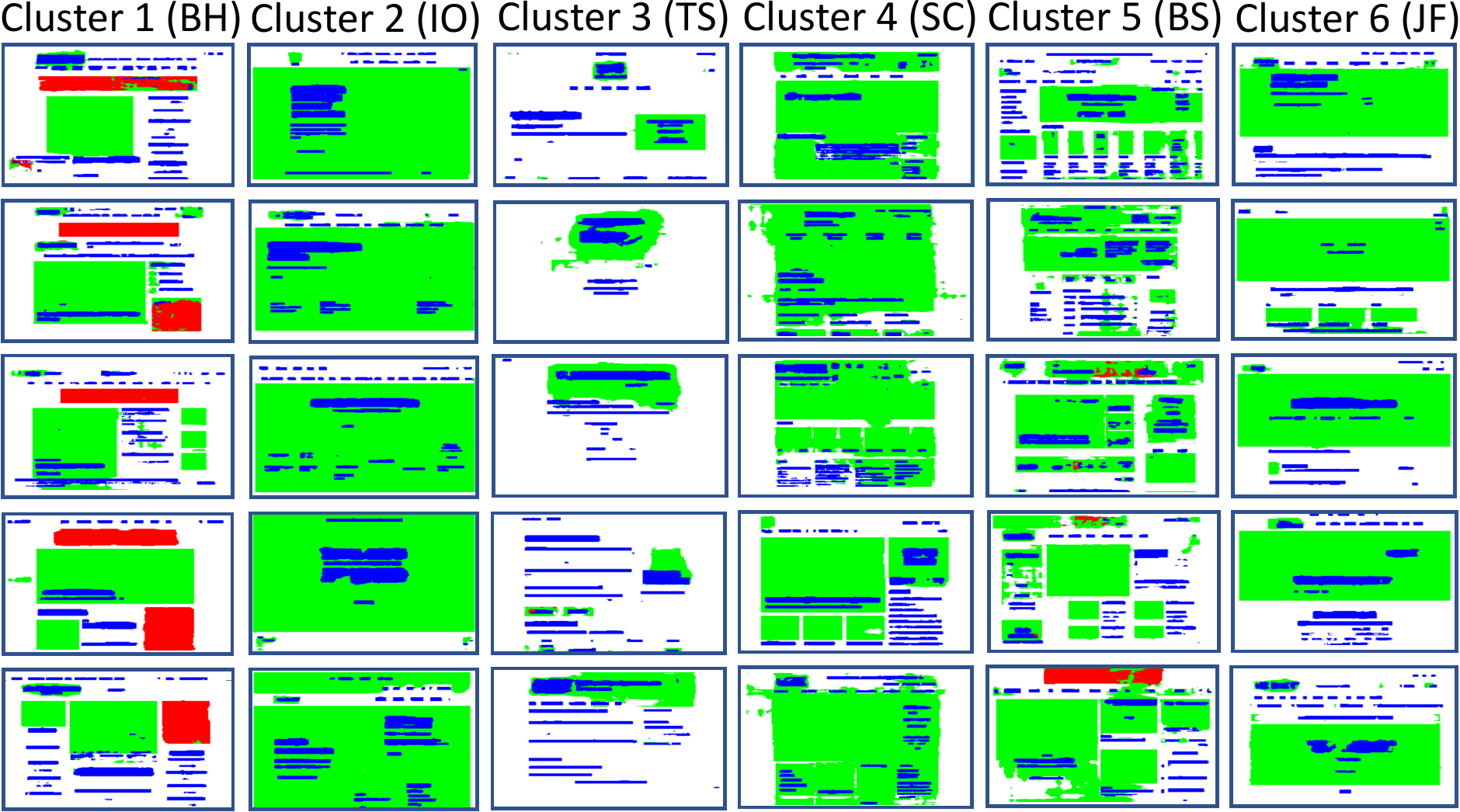}
\caption {Segmentations of webpage images in WebSaliency dataset using PageSegNet on the six different webpage clusters. Here, banner region: red, image region: green, text region: blue, background: white.}
\label{fig:pagesegs}
\end{figure}

We have four semantic categories: images, texts, banners and background (green, blue, red and white in Fig.~\ref{fig:pagesegs}). 
We propose \textit{PageSegNet}, a lightweight version of the Nested UNet model \cite{zhou2018unet++} that segments webpage screenshots from ground-truth layout information,  which we obtain from the page DOM  metadata. On top of PageSegNet, we enforce a pixel-wise categorical cross entropy loss between the predicted and ground truth  segmentations of our 55K images dataset as:
\begin{equation}
L(y^{gt},\hat{y}^{pred}) = - \sum_{j=0}^{M} \sum_{i=0}^{N} y^{gt}_{ij}\log(\hat{y}^{pred}_{ij})
\end{equation}
where the true class is an one-hot encoded vector, $y^{gt}$ and the predicted class is $\hat{y}^{pred}$. $M \times N$ is the image dimension. We adopted a light-weight version of the Nested UNet (UNet++) architecture \cite{zhou2018unet++} for constructing the webpage segmentation network. We resized the webpage input image to size $224\times 224$ before inputting it to the network. The same network architecture was adopted for a model we trained to segment logos in our WebSaliency dataset. We used our PageSegNet network (trained on WebSaliency) to also segment the stimuli of the FiWI dataset~\cite{shen2014webpage}, which only contains webpage screenshots and no layout information. This allowed us to evaluate our model on FiWI; a segmentation of their screenshots allows for a fair comparison.\vspace{1.0mm}\\
\noindent \textbf{Layout representation construction:} We hypothesize that a systematic taxonomy of documents according to layout can assist in the targeted prediction of saliency. To obtain the layout  representations, we first train a UNet model  \cite{ronneberger2015u} on the PageSegNet segmentations, which we call \textit{PageEncoder}. This captures layout information in a 64 dimensional  representation, $R_{Page}$ obtained by down-sampling and flattening the intermediate 2D  representation $R_{Page}'$.
The network is trained using categorical cross entropy loss between the predicted and  reconstructed segmentations. We trained the PageEncoder network using the predicted segmentation map (from PageSegNet) as the input and the reconstructed segmentation as the output. We resized the input image to size $224\times224$ for computational efficiency. For encoding the page layout, we used a light-weight version of UNet \cite{ronneberger2015u}, which provided us a low dimensional representation of the page layout.  Fig.~\ref{fig:segs} depicts the pipeline for layout representation construction. 

\begin{figure*}[h]
\begin{center}
\label{fig:a}\includegraphics[width=16.9cm]{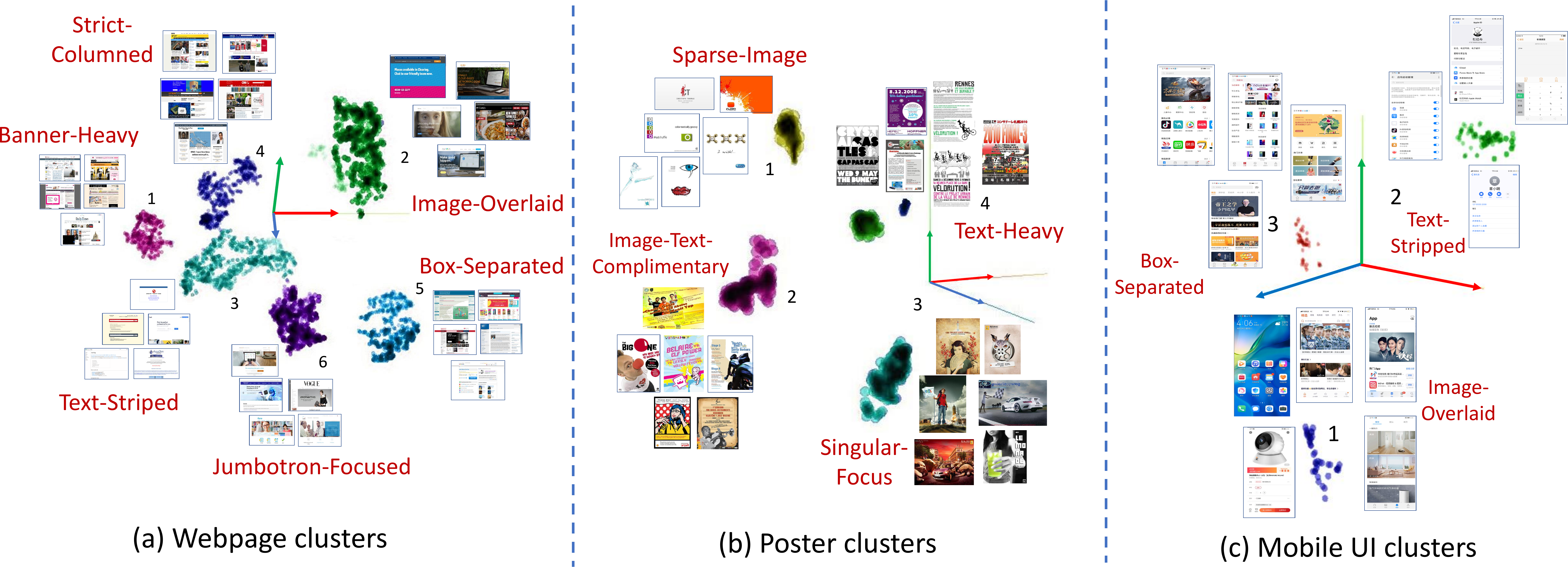}
\end{center}
\caption{t-SNE plot of (a) webpage clusters, (b) poster clusters, (c) mobile UI clusters obtained from our PageEncoder network.}
\label{fig:tsnes}
\end{figure*}

Given a webpage image $I$, we obtain its layout, $L$ and layout representation, $R_{Page}$  using the PageSegNet and the PageEncoder networks respectively. Next, we cluster the PageEncoder layout representations using the K-means++ clustering technique \cite{arthur2007k}. This method assigns the input layout representation, $R^{I}_{Page}$ to the cluster, $c_{min}$ that minimizes the feature distance to its center i.e. $c_{min} = \arg \min_{c} ||R^{I}_{Page} - R^{\mu_c}_{Page} ||_{2}$, where $R^{I}_{Page}$ is the webpage layout representation of input image $I$ and $R^{\mu_c}_{Page}$ is the representation of the center of cluster $c$. While we could have directly clustered the representations of PageSegNet, these are much noisier since PageSegNet encodes more information (high frequency noise in the input images) than just the layout.
To find the optimal number of clusters, we first visualized $R_{Page}$ using t-SNE~\cite{maaten2008visualizing}, which enforces the separation of clusters for interpretation. Given  the t-SNE information, we complete the clustering of the PageEncoder representations using the K-means++ technique. We set K, the number of clusters, to 6, based on the elbow~\cite{apon2006inital} of the within-cluster sum of error plot (across values of K), in combination with  visual inspection of the t-SNE. In Fig.~\ref{fig:tsnes}(a), we show the t-SNE projected to a 3D space. Fig.~\ref{fig:pagesegs} depicts some webpage segmentation instances (from WebSaliency) using PageSegNet of the six webpage clusters. See supplementary for more sample segmentations and the clustering details. 

Table~\ref{tab:optparams} shows the average
Normalized Scanpath Saliency (NSS) scores (see  Sec.~\ref{sec:experiments} for details) of the predicted component FDMs with respect to the ground truth FDM ($FDM_{gt}$) normalized column-wise, with and without using the layout representation for prediction. We see there is a large variability for the component NSS scores across clusters, which is lost in the combined dataset. The low score for banners confirm the concept of banner blindness in webpages \cite{benway1998banner,hervet2011banner}.
\begin{table}[]
\centering

\setlength{\tabcolsep}{2.3pt}
\resizebox{8.6cm}{!}{%
\begin{tabular}{llllllll}
\hline\noalign{\smallskip}
Cluster     & BH & IO & TS & SC & BS & JF & W/o Layout \\
\noalign{\smallskip}
\hline
\noalign{\smallskip}

 $NSS_{Face}$ &  0.193 &  0.145 &  0.158 & \textbf{0.268}  &  0.249 & 0.252  &    0.214    \\ 
$NSS_{Text}$ & 0.256  &  0.231 &  \textbf{0.565} & 0.361  & 0.439  & 0.265   &    0.393    \\ 
$NSS_{Logo}$ & 0.148  & 0.182  & 0.155  & 0.165   & 0.153  & \textbf{0.204}  &    0.184   \\
$NSS_{Banner}$ & \textbf{0.154}  & 0.052  & 0.041  & 0.045   & 0.057  & 0.026  &    0.053   \\
$NSS_{Image}$ & 0.249  & \textbf{0.390}  & 0.075  & 0.160   & 0.101  & 0.262  &    0.156   \\ 
\hline
\end{tabular}
}

\caption{Average
Normalized scanpath saliency scores of the predicted component FDMs, normalized column-wise on WebSaliency. 
}
\label{tab:optparams}
\vspace{-4mm}
\end{table}

\vspace{-2.5mm}
\subsection{Stage 2: Scanpath Prediction}
\label{sec:scanpath_method}
\begin{figure}
\centering
\includegraphics[scale=0.284]{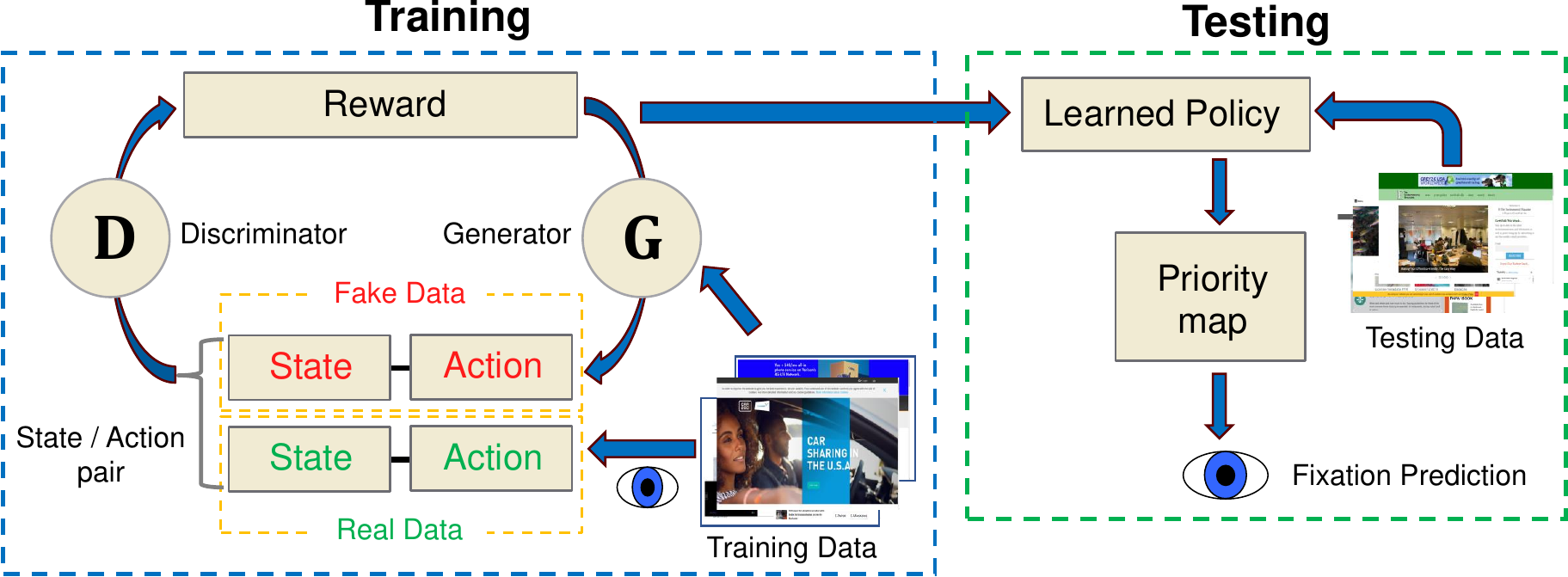}
\caption {Pipeline to predict scanpath on graphic designs using Generative  Adversarial  Imitation Learning. } 
\label{fig:scan_pipe}
\end{figure}

\noindent Having predicted the spatial FDM of a webpage image, we move to the more challenging problem of predicting the spatio-temporal attention scanpath in webpages. We use inverse  reinforcement  learning, recently shown to predict scanpaths during search tasks on natural images~\cite{yang2020predicting}. Specifically, we used Generative Adversarial Imitation
Learning (GAIL)~\cite{ho2016generative} to learn the policy used by people when making fixations during visual search. 
Fig.~\ref{fig:scan_pipe} depicts our GAIL pipeline for predicting scanpaths on graphic design documents. The generator takes an image input and generates a sequence of fixations (fake eye movements), with each movement being an action that is paired with a particular state, mainly the input image, to create a state-action pair. We also have real eye movements in our training dataset, so there is a real state-action pair for the same scene. The discriminator then takes the fake data and the real data as input, and tries to discriminate between the two. Each time the discriminator makes a wrong guess, that particular state-action pair from the generator is rewarded, making it more likely to be generated in the future. During training, the generator therefore becomes increasingly good at fooling the discriminator, meaning that it produces increasingly human-like data that are  difficult to discriminate from real.  

Rather than applying GAIL to search, we modified it for document scanpath prediction by removing the object contingency used by \cite{yang2020predicting} to predict search behavior. Furthermore, \cite{yang2020predicting} used panoptic object segmentations for different natural object categories to construct an internal state representation. Graphic designs however, have a different set of semantic components (text, banners, logos, etc). We thus propose a state representation that reflects the semantic components in a graphic design (the predicted component FDMs from Stage 1) for our IRL model of scanpath prediction. We used the predicted component FDMs and not their segmentation maps because the component FDM predictions are more discriminative than segmentations of only the component locations. More importantly, unlike the dynamic contextual belief maps in \cite{yang2020predicting} that are hand crafted for state space representation, we leverage the automatically learned component FDMs from Stage 1 for predicting scanpaths in Stage 2.

Fig.~\ref{fig:flowchart2} explains the state update rule of our IRL model. We use component FDMs learned by AGD-F and the page layout representation, $R_{Page}'$ learned by PageEncoder (since layout influences attention) for the state representation, which we call Dynamic Component Beliefs. We construct the state representation, $B$ as $B = [Sal_{Face},Sal_{Text},Sal_{Logo},Sal_{Banr.},Sal_{Img.},R_{Page}'].$ Note that $R_{Page}'$ is different from the PageEncoder representation $R_{Page}$ (in Section 3.1.2). The original intermediate 2D feature maps of PageEncoder, $R_{Page}'^{orig}$ consists of 64 channels of size $14\times14$. We obtained a single channel feature map for the layout representation by using a $1\times1$ convolutional layer. Next we resized this single channel map to size $20\times32$ ($R_{Page}'$) in order to match the size of the model state representation. 

To reduce prediction complexity, we spatially discretized every state map into a $20\times32$ grid that defines the action space. Learning a continuous action space would require extensive exploration to obtain the underlying distribution, thus hindering model training.  
Given the IRL data requirements, and the available WebSaliency data, a discretized action space ($20\times32$ grid) helped us constrain the scanpath prediction problem, essentially fixing the spatial resolution of the model's eye movement system. For each fixation, we update the state
by replacing the portion of the low-resolution portion beliefs with the corresponding low-resolution portion at
the new fixation as: $B_0 = L, B_{t+1} = M_t \odot H + (1-M_t) \odot B_t $, where $B_t$ is the belief state after $t$ fixations, $M_t$ is the circular mask generated from the $t^{th}$
fixation, $L$ and $H$ are
the predicted component FDMs obtained using input images at low and high resolutions respectively. The initial state is based on the component beliefs from a low-resolution image, corresponding to a blurred peripheral visual input. For each viewer fixation, we update the state by replacing the low-resolution portion of  the component belief maps with the corresponding high-resolution portion obtained at the new fixation location. This state updating process emulates the shifts of a foveated retina, thus making it useful for modeling human attention behavior. Next, the model generates state-action pairs that emulate human behavior. See supplementary for a visualization of the cumulative foveated retina we use in our model. When generating scanpaths, a fixation location is sampled from the probability map produced by the model and Inhibition-of-Return is applied to prevent revisiting attended locations. 

We followed ~\cite{yang2020predicting} to develop our AGD-S model. AGD-S is composed of three main components: (1) the policy network, (2) the critic network, and (3) the discriminator
network. The policy and critic
networks are jointly updated using the PPO algorithm \cite{schulman2017proximal} by maximizing
the total expected reward, as explained in \cite{yang2020predicting}. 

\vspace{-4mm}
\subsection{Implementation Details}
\textbf{CSPNet}: The dilated inception module consists of a single $1\times1$ convolutional layer followed by three $3\times3$ convolutional layers with dilation rates 4, 8 and 16, the outputs of which are linearly combined to form the encoder output. The encoder network provides the image  representation, $R_{image}$  consisting of 256 maps with spatial resolution $40 \times 60$. We concatenate the page layout representation obtained from PageEncoder, $R_{Page}$ (with 64 channels resized to $40\times60$), with the encoded representation  $R_{image}$ to form the combined representation, $R_{combo}$, which is the input to the dilated inception module. In the decoder, the first two layers have 256 $3 \times 3$ convolutional
kernels with ReLU activation. The prediction  layer has one $3 \times 3$ convolutional kernel with sigmoid activation to down-sample the FDM. For training this model, we used the AdamW optimizer with mini-batch size of 16 images. The learning rate was set to  $10^{-4}$ and scaled down by a factor of 0.1 after every 4 epochs. The network converged within 15 epochs.

\textbf{PageSegNet}: The number of filters used in the convolution layers 1 to 5 (as in the original model) are 16, 24, 32, 48 and 64 respectively. We opted for a smaller architecture with reduced set of filters in order to avoid over-fitting, which provided a well generalized model. We used the AdamW optimizer \cite{loshchilov2017fixing} with an initial learning rate = $10^{-4}$, weight decay = $10^{-4}$ and divided the learning rate by 10 every 15 epochs.

\textbf{PageEncoder}: This network consists of 5 convolutional layers with 16, 24, 32, 48 and 64 feature maps respectively. The \textit{conv5} layer provides us 64 maps, each with spatial dimension $14\times14$. Next, we used a $1\times1$ convolutional layer to reduce the 64 maps into a single spatial map (of size $14
\times14$) that encodes the spatial layout information. We resize this map to size $8\times8$ to further reduce the data dimensionality and flattened the resulting map into a 64-dimemsional vector, which forms our page representation, $R_{Page}$. This network converged faster than our page segmentation network, PageSegNet, using the same training parameters as PageSegNet. 

\textbf{AGD-S}: AGD-S was trained for 15 epochs with an image batch size of 32. It took us approximately 10 minutes to train AGD (for 15 epochs) on a single NVIDIA Tesla V100 GPU. See supplementary for more details on model training.
\vspace{-1.1mm}

\section{WebSaliency Dataset}

\begin{figure}
\centering
\includegraphics[width = 9cm]{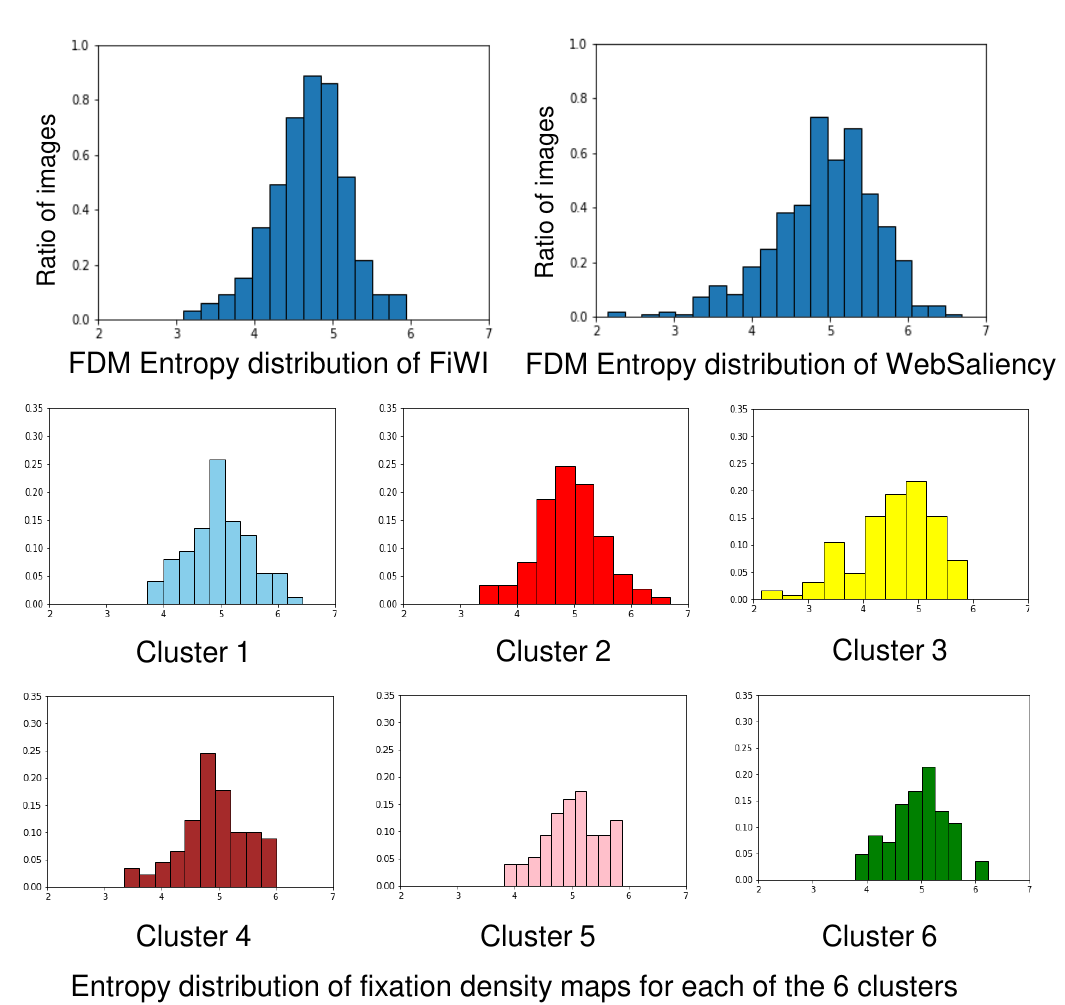}
\caption {Top: Fixation entropy of  FDMs in the WebSaliency and FiWI datasets. Bottom: Fixation entropy of all 6 computed webpage clusters.} 
\label{fig:fixent}
\vspace{-2.5mm}
\end{figure}

\noindent We have collected a large-scale, high-quality gaze dataset of 450 freely viewed webpage images. Each  image  was viewed by 13 or 14 viewers (41 in total). Our dataset is three times larger than the existing FiWI dataset~\cite{shen2014webpage} that has 149 images. FiWI grouped pages manually into 3 broad categories (textual, pictorial and mixed) but is too small to produce other meaningful clusters. Our dataset is large enough that it can be automatically clustered into 6 meaningful categories (Sec~\ref{sec:stage1}), that can also be used for applications such as improving visual search. The dataset is available here:
\href{http://vision.cs.stonybrook.edu/~soura/websaliency.html}{http://vision.cs.stonybrook.edu/$\sim$soura/websaliency.html}
\vspace{1mm}\\
\textbf{Image Sources:} To ensure a uniform distribution of stimuli from across our six clusters, we selected 75 pages per cluster from among our 55k PageSegNet training examples, thus collecting 450 images in total.
\vspace{1mm}\\
\textbf{Gaze collection:} A total of 41 participants (19 females, 22 males; age range 17-23; with normal or corrected-to-normal vision) participated in our data collection. After calibration, participants were instructed to freely view each webpage image for 5 seconds, following \cite{shen2014webpage}. The order of presentation of the web pages was randomized. Eye position was measured using an EyeLink 1000 eye-tracker.
\vspace{1mm}\\
\textbf{Statistics:} We compared the fixation entropy distributions of  WebSaliency and  FiWI 
as in \cite{judd2009learning}. WebSaliency, comprised of six distinct webpage clusters, captures more entropy variation than the existing FiWI. In Fig.~\ref{fig:fixent} we show the fixation entropy distribution of the images in the WebSaliency and the FiWI datasets. Also, we show the fixation entropy distributions of all the computed six webpage clusters. The distribution of FDM entropy widely varies across the 6 clusters, which validates our intuition to create the clusters.  Thus, WebSaliency has sufficient data to improve the overall saliency prediction (Table~\ref{tab: compare_saliency_models_our}). See supplementary for details.

\section{Experiments}
\label{sec:experiments}

\subsection{Evaluation of the Saliency Prediction Model}
\label{sec:evaluation}
\noindent We compare our AGD-F model with natural saliency models: Itti \cite{itti1998model}, Deep Gaze II \cite{kummerer2016deepgaze}, SalGAN \cite{pan2017salgan}, DVA \cite{wang2017deep}, SAM-ResNet \cite{cornia2018predicting}, UAVDSM \cite{he2019understanding}, DI Net \cite{yang2019dilated}, EML-NET \cite{jia2020eml}, webpage saliency models: MKL \cite{shen2014webpage}, MMF \cite{li2016webpage} and SPPL \cite{gu2019element}, TaskWebSal (FreeView) \cite{zheng2018task} and the graphic design saliency model, UMSI \cite{fosco2020predicting}. We quantitatively compare  over the metrics: Normalized Scanpath Saliency (NSS), Cross Correlation (CC), KL divergence (KL), the Judd AUC (AUC-J) \cite{bylinskii2018different} and the Shuffled AUC (sAUC). We randomly selected 58 images from WebSaliency for testing and the rest for training.  

\begin{figure*}[h]
\centering
\includegraphics[width = 18cm]{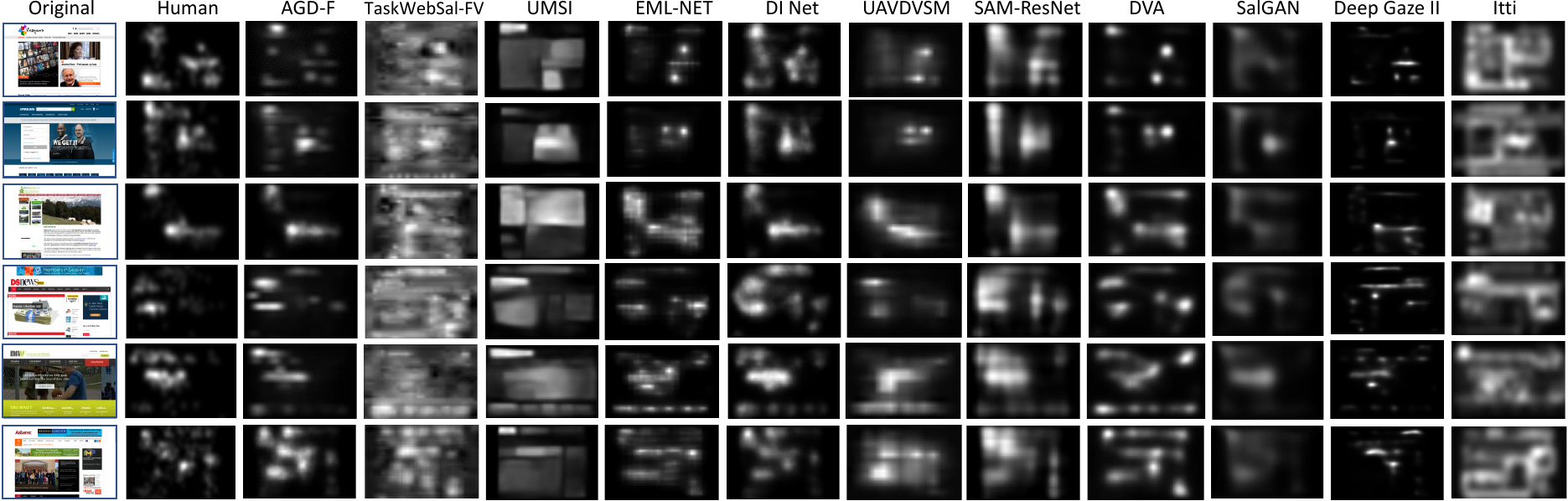}
\caption {Qualitative evaluation of how human attention compared to different methods (Left to Right): AGD-F, task-free (free viewing) saliency map from \cite{zheng2018task}, UMSI importance \cite{fosco2020predicting}, EML-NET \cite{jia2020eml}, DI Net \cite{yang2019dilated} (trained-WebSaliency), UAVDVSM \cite{he2019understanding}, SAM-ResNet (trained-WebSaliency) \cite{cornia2018predicting}, Deep Visual Attention (DVA) \cite{wang2017deep}, SalGAN \cite{pan2017salgan}, Deep Gaze II \cite{kummerer2016deepgaze}, Itti \cite{itti1998model}  on our WebSaliency dataset.} 
\label{fig:quals_ourdata}
\vspace{-2.2mm}
\end{figure*}

Fig.~\ref{fig:quals_ourdata} shows the predicted FDMs of the compared methods for six test webpage instances. Qualitatively, natural saliency models UAVDVSM, EML-NET and Deep Gaze II  produce inaccurate text saliency predictions (rows 2 and 5, where logo saliency is not predicted)  and fail to capture the F-bias (the F-shaped pattern of eye movement) in row 2. Also, SAM-ResNet and Deep Gaze II inaccurately allocate high saliency to certain regions, e.g. in the right column of images in row 4. Similar inconsistencies can be observed in the predictions using the EML-NET and the DVA models. For example, in rows 3 and 5, these models overestimate the saliency of the text snippets on the top right and bottom image regions respectively. We also show the task-free FDMs obtained using the TaskWebSal model and the importance maps using the UMSI model on the same image instances. We see that the FDM obtained from the task-free subnet of the TaskWebSal model is very diffused and inaccurate compared to the ground truth. This could be attributed to the fact that the model has only been trained on 200 webpage images on task-specific attention data. On the other hand, importance maps predicted using the UMSI model on WebSaliency images do not well predict the ground truth eye fixation data we collected, as shown in Table III and Fig.~\ref{fig:quals_ourdata}. We attribute this poor predictive performance to the fact that importance maps predict similar attention values for all regions within a graphic design component, e.g. the logos and image component regions in Fig.~\ref{fig:quals_ourdata}. Also observed in Fig.~\ref{fig:quals_ourdata}, the predicted importance map highlights the regions that are not salient to humans, e.g in row 3 the Human FDM (column 2) indicates that the image blocks are not salient to the viewers, but the importance map predicts otherwise (column 5). Similar observations could be made for other instances. The IMP1K dataset \cite{fosco2020predicting} (which the UMSI model has been trained on) calculates importance maps based on region importance ratings from Amazon Mechanical Turk viewers.  Note that importance maps and fixation-density maps are qualitatively different, with our experiments suggesting that eye fixations are more informative of where people allocate attention compared to importance maps. The predictions of DI Net trained on WebSaliency are visually closer to our predictions, although the maps are less compact and inaccurate compared to our FDMs, especially in rows 2 and 4. Our model produces more accurate and compact fixation density maps by better predicting the saliency of the different webpage components compared to other methods. For example, the text saliency in rows 3 and 5 and the saliency of the faces in rows 1 and 2 are better predicted than the compared methods. In the last row of Fig.~\ref{fig:quals_ourdata} we also depict an instance where the proposed model overestimates the saliency of the banners and the people (standing) in the image block, while failing to highlight the increased saliency of the upper textual regions in the text block  accompanying the image block. This could be due to the fact that the upper textual region is more  centrally located compared to other salient components, and therefore attracted more attention within the first 5 seconds. Aggregating attention data over higher duration might help reduce the gap between the human FDM and the model prediction.

Tables~\ref{tab: compare_saliency_models_fiwi} and ~\ref{tab: compare_saliency_models_our} list different performance metrics for the compared methods.  As seen, AGD-F  outperforms previous methods on all metrics on both FiWI and WebSaliency. We did not compute KL divergence on FiWI (Table ~\ref{tab: compare_saliency_models_fiwi}) and did not evaluate MKL, MFF, SPPL on WebSaliency (Table ~\ref{tab: compare_saliency_models_our}) due to lack of publicly available code.

\begin{figure*}
\centering
\includegraphics[width =16.5cm]{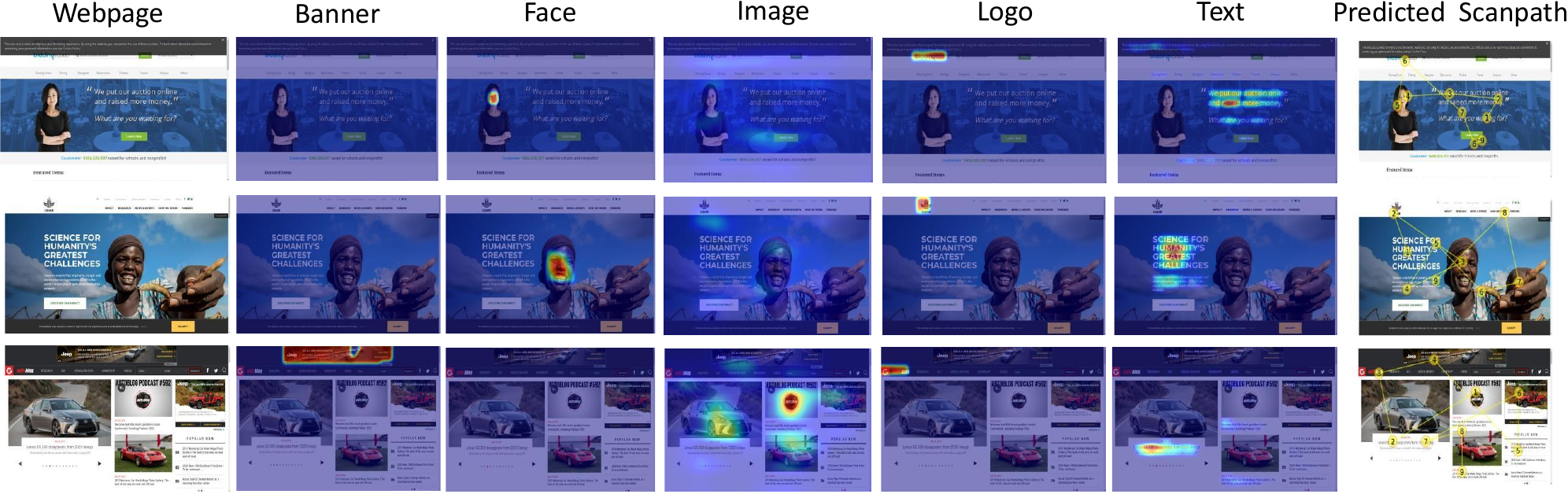}
\caption {The normalized predicted component fixation density maps used by our AGD-S model for predicting scanpaths on webpage images.}
\label{fig:components}
\vspace{-4mm}
\end{figure*}

\begin{table}
\centering
\resizebox{9.3cm}{!}{%
\begin{tabular}{lM{1.1cm}lM{1.1cm}lM{1.1cm}lM{1.1cm}l}
\hline\noalign{\smallskip}
\textbf{Method}    & \textbf{NSS $\uparrow$} & \textbf{CC $\uparrow$} & \textbf{AUC-J $\uparrow$} & \textbf{sAUC $\uparrow$}\\
\noalign{\smallskip}
\hline
\noalign{\smallskip}
Itti  & 0.617  & 0.314 & 0.690  & 0.629 \\   
Deep Gaze II  & 1.229  & 0.488 & 0.797 &  0.625 \\ 
SalGAN  & 1.195  & 0.598 & 0.785 &  0.627 \\
DVA  & 1.215  &  0.542 & 0.787 &  0.687 \\
SAM-ResNet  & 1.246 & 0.595 & 0.791 &   0.673\\ 
UAVDVSM  & 1.058  &  0.509  & 0.747 &  0.642 \\
UMSI &  0.938 & 0.457 & 0.755 & 0.675 \\
TaskWebSal-FreeView &  0.893 & 0.425 & 0.736 & 0.677\\
MKL  & 0.881 & 0.429 & - &  0.720\\
MFF  & 0.905 & 0.441 & - &   0.731\\
SPPL  & 1.085 & 0.547 & - &   0.751\\
AGD-F   (W/o-L FiWI)  & 1.606 & 0.735 & 0.767 &  0.748\\ 
AGD-F (W/o-L WebSaliency)  & 1.639 & 0.766 & 0.843 &   0.753\\ 
AGD-F (W-L WebSaliency)  & \textbf{1.662} & \textbf{0.783} & \textbf{0.851} &  \textbf{0.758} \\
\hline
\end{tabular}
}
\caption{Performance of saliency prediction models on 28 test images of FiWI. W/o-L FiWI - Without layout representation trained on FiWI, W-L WebSaliency - With layout represenatation  trained on WebSaliency.}
\label{tab: compare_saliency_models_fiwi}
\vspace{-3.0mm}
\end{table}

\begin{table}
\hfill
\centering
 \resizebox{9.0cm}{!}{%
\begin{tabular}{llllll}
\hline\noalign{\smallskip}
\textbf{Method}                                & \textbf{NSS $\uparrow$} & \textbf{CC $\uparrow$} & \textbf{KL $\downarrow$}
& \textbf{AUC-J $\uparrow$}  & \textbf{sAUC $\uparrow$}\\ \hline
\noalign{\smallskip}
Itti  & 0.769 & 0.367 & 0.840 & 0.710 &  0.661\\
Deep Gaze II  & 1.380  & 0.574 & 3.449 & 0.815 & 0.644\\
SalGAN (Tr. WebSaliency) & 1.458  & 0.637 & 0.622 & 0.818 & 0.703\\
DVA & 1.260  & 0.571 & 0.701 & 0.805 & 0.711\\
UAVDVSM  & 1.133 & 0.519 & 0.858 & 0.739 &  0.668\\
SAM-ResNet  & 1.284 & 0.596 & 1.506 & 0.795 & 0.717\\
EML-NET  & 1.277 & 0.565 & 2.11 & 0.790 & 0.702\\
UMSI &  1.042 & 0.444 & 1.335 & 0.757 & 0.698\\
TaskWebSal-FreeView &  1.107 & 0.525 & 0.784 & 0.769 & 0.714\\
SAM-ResNet (Tr. WebSaliency) &  1.532 & 0.718 & 0.994 & 0.828 & 0.725\\
DI Net (Tr. WebSaliency) & 1.777 & 0.798 & 0.690 & 0.852 & 0.739\\   
AGD-F (W/o-L)  & 1.802 & 0.815 & 0.637 & 0.858 & 0.753\\  
AGD-F (W-L)  & \textbf{1.821} & \textbf{0.823} & \textbf{0.619} & \textbf{0.865} & \textbf{0.762}\\
\hline
\end{tabular}
}
\caption{Performance of saliency models on 58 test images of WebSaliency. W/o-L - Without layout, W-L - With layout representation trained on WebSaliency.}
\label{tab: compare_saliency_models_our}
\vspace{-3.3mm}
\end{table}
\vspace{2mm}
\noindent\textbf{Ablation Study}:
\noindent \subsubsection{Importance of  layout-based feature representation}
\label{sec:rep_imp}
We consider a naive alternate representation of our segmentation map. We define $R_{img}$ as the ratio of the number of image pixels segmented to the total number of pixels in the stimuli. We computed the corresponding ratios for every component and construct a vector $SegStats = [R_{img},R_{text},R_{banner},R_{bg}]$, which was used for clustering to obtain the saliency maps. Table~\ref{tab: pageenc}  compares the performance metrics of the two representations and  shows that the PageEncoder representation improves prediction accuracy.

\begin{table}
\centering
\resizebox{8.57cm}{!}{%
\begin{tabular}{llllll}
\hline\noalign{\smallskip}
\textbf{Method}                                & \textbf{NSS $\uparrow$} & \textbf{CC $\uparrow$} & \textbf{KL $\downarrow$}
& \textbf{AUC-J $\uparrow$}  & \textbf{sAUC $\uparrow$}\\ \hline
\noalign{\smallskip}
AGD-F (Rep-PageEnc)  &  \textbf{1.821} & \textbf{0.823} & \textbf{0.619} & \textbf{0.865} & 0.762 \\ \vspace{1mm}
AGD-F (Rep-SegStats)  &    1.752                       &      0.810       &        0.672    &  0.843 & 0.748 \\
\hline
\end{tabular}
}

\caption{Evaluation using PageEncoder representations (Rep-PageEnc) of AGD-F vs. feature vectors constructed using segmentation statistics (Rep-SegStats) without the layout representation.}
\label{tab: pageenc}
\vspace{-6mm}
\end{table}
\subsubsection{Contribution of the salient webpage components} Table \ref{tab: pathway_contribution} compares the contributions of the salient page components to the final performance. We see that textual  regions attract the most attention. Texts, logos and faces are the most influential components in attracting attention. Banners are the least  predictive of the allocated attention, which could be attributed to the banner blindness phenomenon while viewing webpages, which further supports our findings in Table~\ref{tab:optparams}. However, considering all the components together leads to the best prediction performance.

\begin{table}
\centering
\resizebox{9cm}{!}{%
\begin{tabular}{llllll}
\hline\noalign{\smallskip}
\textbf{Method}    & \textbf{NSS $\uparrow$} & \textbf{CC $\uparrow$}
& \textbf{KL $\downarrow$} & \textbf{AUC-J $\uparrow$}  & \textbf{sAUC $\uparrow$}\\
\noalign{\smallskip}
\hline
Face (W/o-L)  & 0.484 & 0.236  & 15.73  &  0.535  & 0.524\\
Face  & 0.492   &  0.251    &   13.54  &  0.538  & 0.526\\
Text (W/o-L)  &   1.718 &  0.686   &  3.610 &  0.765 & 0.702 \\
Text   &  1.732  &  0.702   &  3.312 &  0.781  & 0.707 \\ 
Logo (W/o-L)  &   0.508 &  0.172   & 20.95  & 0.550 & 0.538 \\
Logo   &  0.546  &  0.176  &  20.23 &  0.557  & 0.540 \\
Banner (W/o-L)  &   0.128 &  0.025   & 19.80 & 0.504 & 0.518 \\
Banner   &  0.126  &  0.025  &  19.87 &  0.504  & 0.518 \\
Image (W/o-L)  &   0.675 &  0.347   & 12.15  & 0.607 & 0.566 \\
Image  &  0.694  &  0.351  &  12.01 &  0.612  & 0.569 \\
Face + Text (W/o-L) & 1.741  &  0.709 &  2.940  & 0.787 & 0.709\\
Face + Text &   1.752  &  0.725 &  2.820  & 0.798  & 0.717\\
Face + Text + Logo (W/o-L)  &   1.784  &  0.741 &  1.933 &  0.812 & 0.720\\
Face + Text + Logo  &  1.791 & 0.762 &  1.754 &  0.823 & 0.724 \\ 
Face + Text + Logo + Image (W/o-L)  &   1.801  &  0.814 &  0.640 &  0.856 & 0.753\\
Face + Text + Logo + Image  &  1.821 & 0.822 &  0.620 &  0.865 & 0.762 \\
AGD-F (W/o-L)  & 1.802 & 0.815 & 0.637 & 0.858 & 0.753 \\ 
AGD-F  & \textbf{1.821} & \textbf{0.823} & \textbf{0.619} & \textbf{0.865}& \textbf{0.762} \\
\hline
\end{tabular}
}
\caption{Performance evaluation of  the different salient webpage components in AGD-F on our  WebSaliency dataset.}
\label{tab: pathway_contribution}
\vspace{-1mm}
\end{table}

\noindent \subsubsection{Comparison of computational cost of different saliency prediction models}
\label{sec:compcost}
\begin{table}
\hfill
\centering
 \resizebox{8.5cm}{!}{
\begin{tabular}{llll}
\hline\noalign{\smallskip}
\textbf{Method}              & \textbf{Input Size}                  & \textbf{ Parameters (M)} & \textbf{GFLOPs}\\ 
\hline
\noalign{\smallskip}
SalGAN & $192\times256$ & 31.79  & 91.46\\
Deep Gaze II & $192\times256$ & 20.44  & 20.22\\
UAVDVSM  & $224\times224$ & 14.72 & 30.76\\
SAM-ResNet  & $192\times256$ & 70.09 & 72.91\\
DI Net & $320\times480$  & 27.04 & 15.65\\
EML-NET  & $192\times256$ & 47.09 & 16.24\\
UMSI & $240\times320$ &  29.92 & 0.12\\
TaskWebSal & $224\times224$ &  136.54 & 165.83\\
AGD-F (W/o-L)  & $320\times480$ & 32.95 & 16.55\\ 
AGD-F (W-L)  & $320\times480$ & 34.08 & 17.02\\
\hline
\end{tabular}
}
\caption{Computational details (total number of model parameters and FLOPs) of different saliency prediction models.}
\label{tab: computation_cost}
\vspace{-4.6mm}
\end{table}
 
In Table~\ref{tab: computation_cost} we report  the number of model parameters and GFLOPs of the compared deep learning models to analyze their computational requirements. Our model has significantly fewer parameters and GLOPs compared to the natural saliency model  SAM-ResNet \cite{cornia2018predicting} and the webpage saliency model TaskWebSal \cite{zheng2018task}. The number of model parameters is similar to UMSI \cite{fosco2020predicting}, the state-of-the-art for predicting importance on graphic designs, although UMSI uses  significantly less GFLOPs for prediction. Also, the number of GFLOPs and model parameters compare well with other natural saliency models such as DI Net, EML-NET and Deep Gaze II. Thus, our model has reasonable computational requirements  despite the use of multiple sub-networks. We attribute this to  the fact that the PageSegNet and PageEncoder sub-networks are lightweight versions of the U-Net++ model \cite{zhou2018unet++} and our CSPNet model uses dilated convolutions, known  to significantly reduce computational cost compared to other encoder-decoder models for saliency predictions (e.g. Deep Gaze II, SalGAN, UAVDVSM, etc.) that use convolutional networks with different kernel sizes.  
\vspace{-2.5mm}
\subsection{Evaluation of the Scanpath Prediction Model}
\begin{figure*}[h]
\centering
\includegraphics[width=18.2cm]{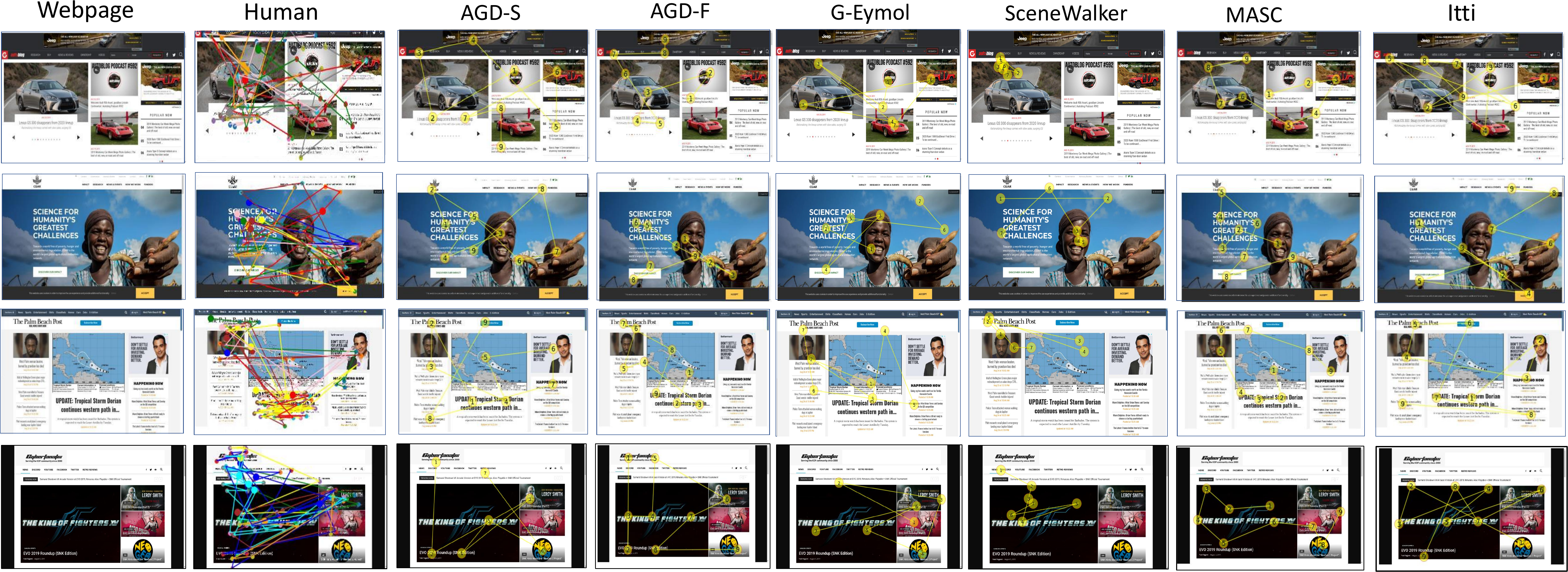}
\caption {Sample scanpaths generated by the compared models on instances from our WebSaliency dataset. Please view in magnification.}
\label{fig:scanpaths}
\vspace{-3.2mm}
\end{figure*}

\begin{table}
\centering
\hfill
\centering
\resizebox{9cm}{!}{%
\begin{tabular}{llllll}
\hline\noalign{\smallskip}
 & & & \textbf{MultiMatch} \\
\hline\noalign{\smallskip}
\textbf{Method}    &
\textbf{Sequence Score}    & \textbf{Shape} & \textbf{Direction}
& \textbf{Length} & \textbf{Position} \\
\noalign{\smallskip}
\hline
Itti  &  0.177         &         0.781    &        0.676     &  0.778  & 0.594\\
MASC  &    0.169  &  0.788   &  0.580   &  0.818  & 0.514\\
SceneWalker  &   0.194   &  \textbf{0.843}  &   0.616 &  \textbf{0.842}  &  0.562\\
G-Eymol &   0.218   &  0.820  &   0.673 &  0.816  &  0.681\\
AGD-F (w layout)      &    0.203   &      0.787       &        0.642  &  0.771 & 0.677  \\

AGD-S (w/o layout)       &    0.221     &      0.814       &        0.663    &  0.805  & 0.698\\
AGD-S (w layout)       &    \textbf{0.224}   &    0.820       &        \textbf{0.677}    &  0.813  &  \textbf{0.708}\\
Inter-Observer &    0.314       &      0.880       &        0.761    &  0.886  & 0.775  \\
\hline
\end{tabular}
}
\caption{Quantitative evaluation of the compared methods for scanpath prediction on our WebSaliency test dataset.}
\label{tab: quant_scanpath}
\vspace{-6mm}
\end{table}

\noindent We  compare  our  AGD-S  model with  the following baselines: (1) \textbf{Itti} model~\cite{itti1998model}, (2) \textbf{AGD-F} (our saliency model) with Winner-Take-All (WTA) and Inhibition of Return (IoR) applied on the FDM, (3) the \textbf{MASC} model \cite{adeli2017model} for scanpath prediction, (4) the \textbf{SceneWalker} model \cite{schwetlick2020modeling}, and (5) the \textbf{G-Eymol} model \cite{8730418}. We compared the models using: (1) \textbf{Sequence
Score (SS)} - by converting the predicted and the human scanpaths into corresponding strings of fixation
cluster IDs and then measuring string similarity using \cite{needleman1970general}, and (2) \textbf{MultiMatch} \cite{dewhurst2012depends}, which considers shape, direction, length, and position aspects of scanpath similarity. We evaluated the first seven fixations on the same test set as our saliency model evaluation. In Table~\ref{tab: quant_scanpath} we  compare prediction performance as well as inter-observer (IO) performance, where one subject's scanpath  predicts  another subject's scanpath for
the same image. High Sequence Score and MultiMatch values suggest a high agreement in how different people viewed these documents, despite the potential for idiosyncratic differences. As seen in Table~\ref{tab: quant_scanpath}, scanpaths predicted by our model are more similar to those of people compared to baselines. We also compare our AGD-S model without layout information in the state representation, where we found adding layout information improves the prediction performance.

\begin{figure}
\centering
\includegraphics[width =7.6cm,height=10.8cm]{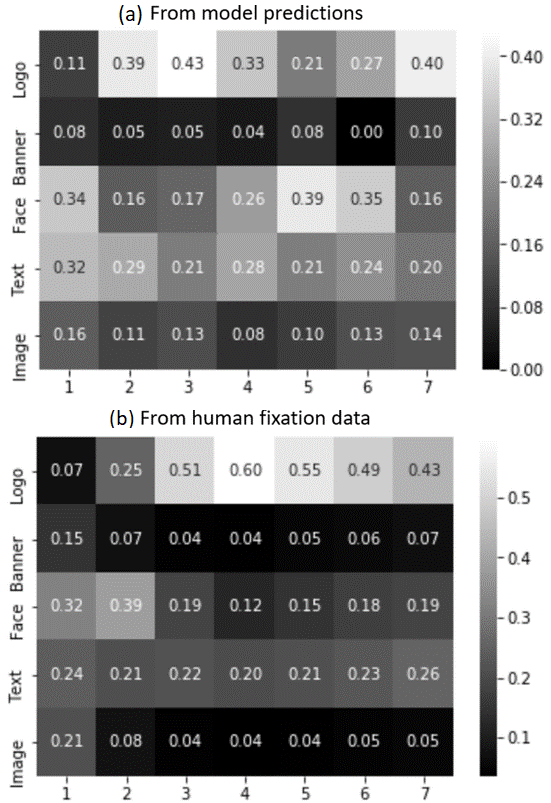}
\caption{(a) Average reward value attributed by our AGD-S model to the different webpage components, (b) proportion of human fixations landing on the different webpage components on our WebSaliency dataset, divided by the component area and normalized by the webpage components (column-wise) across the first seven fixations.}
\label{fig:priority_dist}
\vspace{-2mm}
\end{figure}

In Fig.~\ref{fig:components} we show the normalized predicted component FDMs, used as the state representation for our AGD-S model of scanpath, overlaid on three webpage images from our test dataset
. Here we visualize how the different predicted component FDMs (banners, faces, texts, logos, and images) contribute to the predicted scanpath. Observe that webpage components having high values in their corresponding FDMs are attended more frequently in the predicted scanpaths. In Fig.~\ref{fig:scanpaths}, we show sample scanpaths predicted using the compared methods. AGD-S predicts scanpaths more closely resembling human scanpaths than other baselines. The other models often predicted that the fixations in a scanpath would be to widely separated webpage locations (e.g., a movement from one corner of the image to the other), and these predicted scanpaths often missed the important content of the wedpage, based on human ground truth. 
However, we see that the AGD-S model and the viewers both tended to look at the logos with fixations 2 and 3, also seen in Fig.~\ref{fig:priority_dist}(a), where the logo reward peaks between fixations 2 to 4. Also, the first new fixations during web viewing tended to land on an image or text region. We also show in row 4 of Fig.~\ref{fig:scanpaths} an instance where the predicted scanpath using our AGD-S method does not align well with the human scanpaths.

In Fig.~\ref{fig:priority_dist}(a), we plot the average reward value attributed by our model (normalized by component area column-wise) to different webpage components across the first seven fixations. In Fig.~\ref{fig:priority_dist}(b), we plot the proportion of human fixations landing on the different webpage components on our WebSaliency dataset (normalized by component area column-wise). Figures~\ref{fig:priority_dist}(a) and (b) show that the relative priorities of the different webpage components are consistent across the AGD-S model predictions and the human fixations
, which explains the faithful prediction of attention allocation on webpage images using our model. Our analysis also suggests that logos in a webpage are highly successful in attracting attention, followed by faces and texts. Our analysis also confirms \textit{banner blindness}, a phenomenon where viewers tend to ignore banner-like information, here indicated by low average reward values. 
\vspace{-7.0mm}
\subsection{Dwell map}

\begin{figure*}
\centering
\includegraphics[width =14.9cm,height = 6cm]{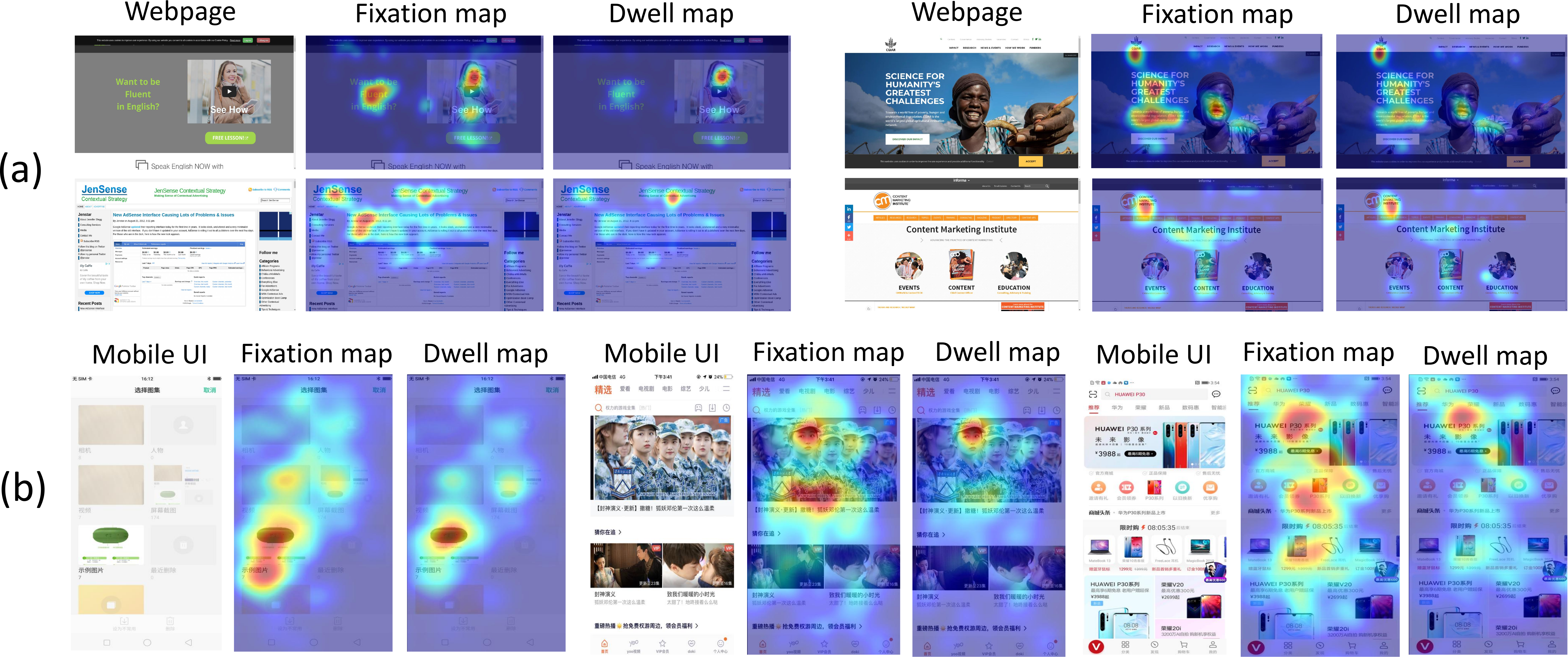}
\caption {The fixation density maps and the dwell maps for (a) webpage images and (b) Mobile UI images.}
\label{fig:dwell_row}
\end{figure*}

The above experiments analyze the spatial FDM  and the temporal order of the fixations while viewing a document, but human viewing behavior also varies in the duration of these fixations. To account for the fixation duration, we simply add the duration value at each fixated location to compute a dwell map, which highlights image regions that have been dwelled upon longer during viewing. We computed the dwell density map as follows:

\begin{equation}
DM_{Dwell} = G_{\sigma}\sum_{x,y}d(x,y)*\mathbbm{1}(x,y)
\end{equation}

\noindent where, $\mathbbm{1}$ is the indicator function that takes value 1 at the fixated pixel location $(x,y)$, $d$ is the duration of the fixation at that location and $G_{\sigma}$ is a Gaussian blur applied on the dwell map to obtain the dwell density map $DM_{Dwell}$. 

\begin{figure}
\centering
\includegraphics[width =7.2cm]{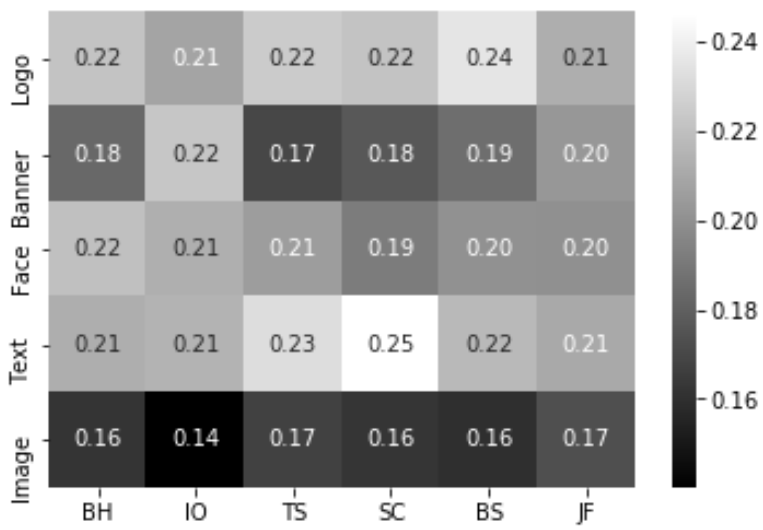}
\caption {Column and component area-normalized (across the five webpage components) fixation-dwell map difference for all webpage layouts.}
\label{fig:dwell_col_norm}
\vspace{-2.2mm}
\end{figure}

Fig.~\ref{fig:dwell_row} shows a qualitative comparison of fixation density maps and dwell maps for two different types of graphic designs, (a) webpages (b) and mobile UIs. The dwell maps highlight the time people spend viewing different regions. As the examples from Fig.~\ref{fig:dwell_row}(a) and (b) show, different document components, such as faces, texts and objects, receive different intensity values in the fixation-density and the dwell-density maps. Given the related but different prioritization performed by these two kinds of maps, we replicated some analyses on the dwell-intensity maps for comparison. Fig.~\ref{fig:dwell_row} visualizes some map comparisons. Fig.~\ref{fig:dwell_row}(a), first row and left column, shows faces being more prioritized in the dwell maps compared to fixation maps. A similarity selectivity exists for text, where some text is dwelled upon more than others (row 2, left column), and for webpage logos (row 2, right column). Similarly, in Fig.~\ref{fig:dwell_row}(b), we see preferences for faces (row 1, middle column) and certain pictorial regions (row 1, left) in the dwell maps computed for mobile UI. 
In Fig.~\ref{fig:dwell_col_norm} we compute a difference between the fixation-density and dwell maps, where mean intensity codes the pixel-wise absolute value of the difference, obtained for different webpage layouts  normalized across the webpage components. Consistent with the previous observations, we see that differences between dwell maps and fixation maps are greatest for text and logos, perhaps suggesting a difference in viewer engagement on these region types that is captured by dwell maps.

\vspace{-2.0mm}
\section{Model Generalization}
\label{sec:generalization}
\vspace{-3mm}
\begin{figure}[h]
\centering
\includegraphics[width=7.9cm]{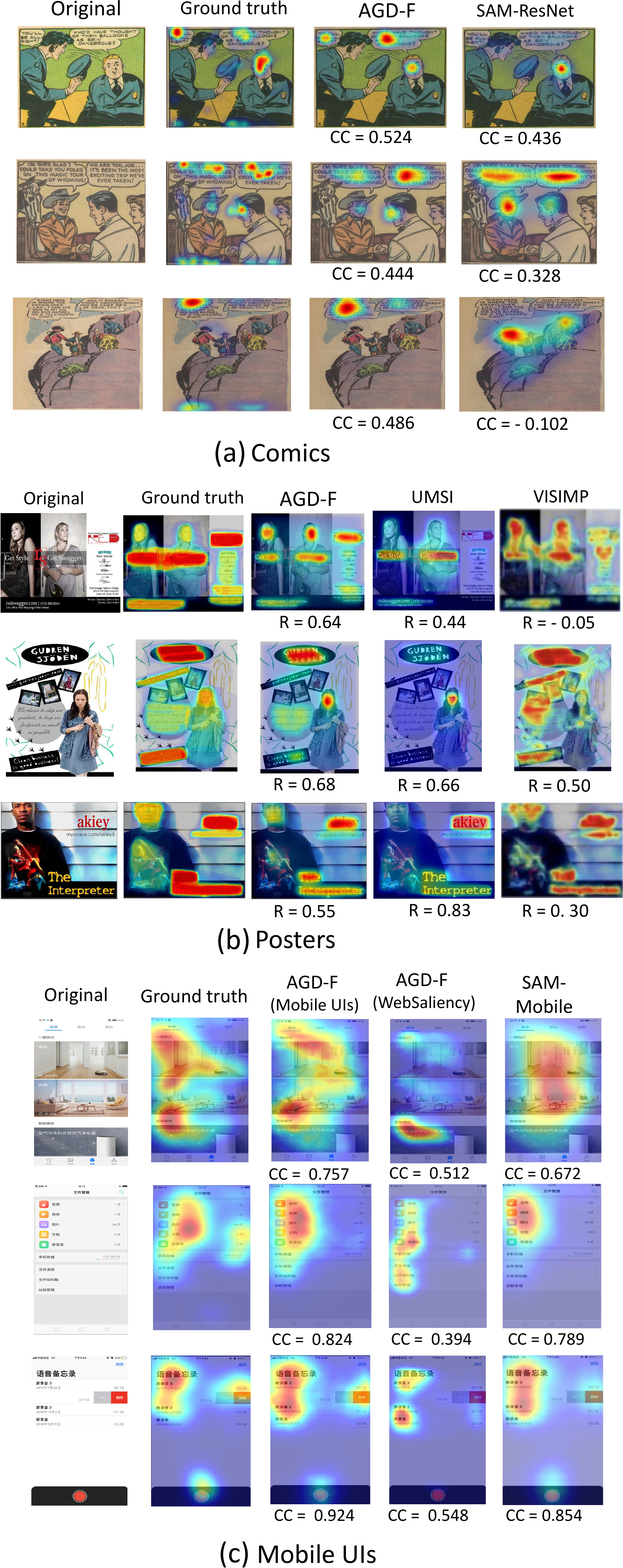}
\caption {\textbf{(a):} Saliency detection on comics in DeepComics~\cite{bannier2018deepcomics}, (CC is Cross Correlation score) \textbf{(b):} Saliency detection on poster images in GDI \cite{bylinskii2017learning} (R is  Spearman correlation coefficient). \textbf{(c):} Saliency detection on mobile UI images in \cite{leiva2020understanding} (CC is Cross Correlation score).} 
\label{fig:joint_designs}
\vspace{-2mm}
\end{figure}

\begin{figure}
\centering
\includegraphics[width=9cm]{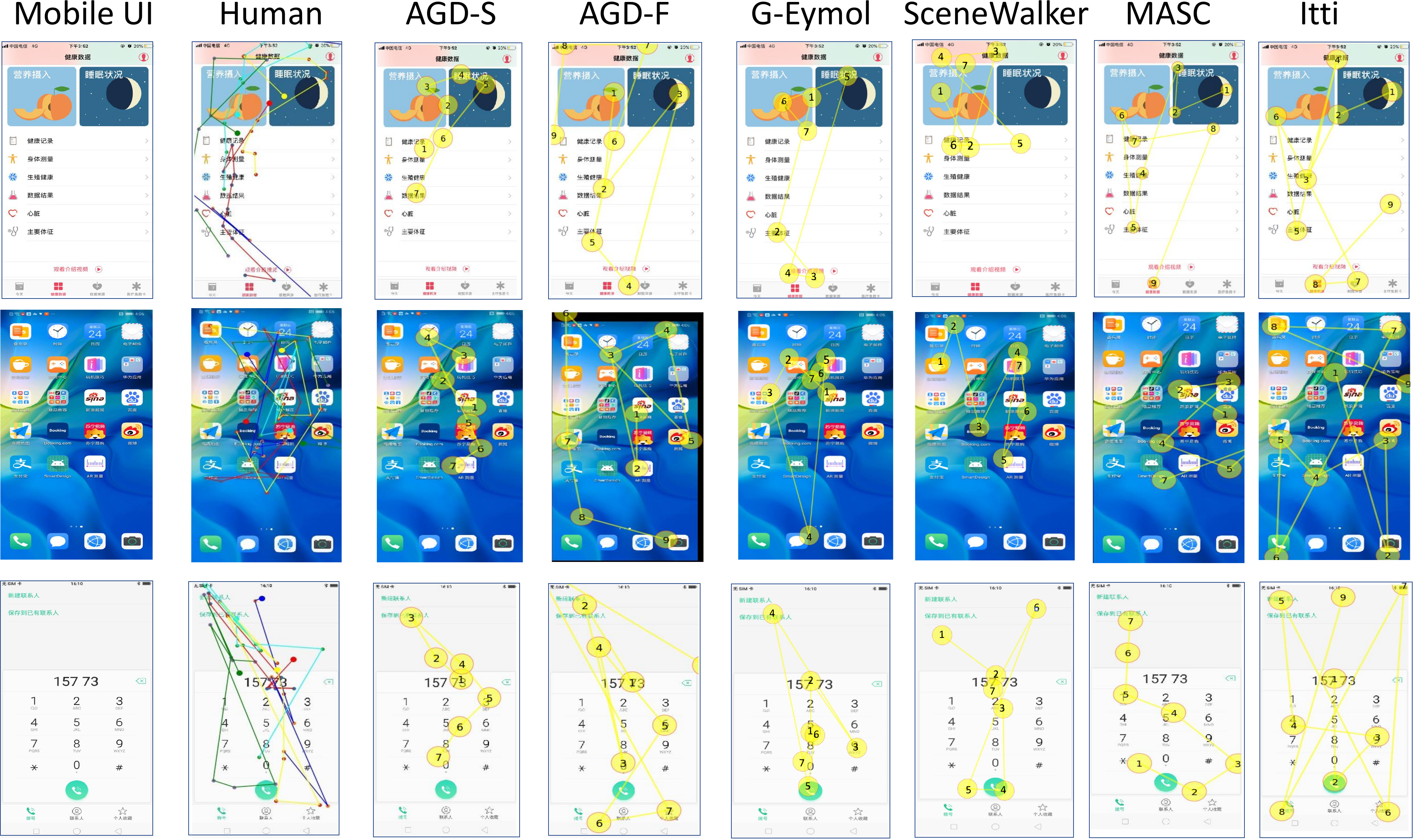}
\caption {Sample scanpaths generated by the compared models on instances from the mobile UI dataset \cite{leiva2020understanding}. Please view in magnification.}
\label{fig:scanpaths_mob}
\end{figure}
\vspace{-2.0mm}
\noindent\subsection{Saliency in Comics:} Using the DeepComics saliency dataset \cite{bannier2018deepcomics}, we found that the segmentations of our Image-Overlaid webpage cluster closely resemble that of comics images. We applied the same model trained on WebSaliency for saliency detection in comics.
Fig.~\ref{fig:joint_designs}(a) depicts our saliency predictions on comics images in DeepComics.  AGD-F more faithfully detects the salient face and text regions as compared to SAM-ResNet. On this dataset, we obtained mean CC score = 0.653, versus  SAM-ResNet's 0.609. AGD-F outperforms SAM-ResNet on DeepComics. In row 1, the SAM-ResNet method inaccurately assigns a high saliency value to the rightmost face only. Our method, however, also detects the upper text region to be salient resulting in more accurate prediction. In rows 2 and 3, the saliency of the faces and textual regions are more accurately predicted by our model respectively compared to SAM-ResNet. We used the same trained model on our WebSaliency dataset for saliency prediction of comic images.  Detailed cluster analysis wasn't feasible due to the limited collection of comic images (23 only) \cite{bannier2018deepcomics}. Therefore, we used our model without layout representation to obtain the predictions. More visualizations in the supplementary.
\vspace{-2.5mm}
\noindent\subsection{Saliency in Posters:} 
\vspace{-1mm}
We re-trained our model AGD-F on the Graphics Design Interface (GDI) dataset of VisualImportance (VISIMP)~\cite{bylinskii2017learning}. We retrained AGD-F as, in VISIMP \cite{bylinskii2017learning}, the ground truth importance maps were obtained by averaging important region annotations from MTurk participants, and thus are different from FDMs that are obtained using an eye tracker. Fig.~\ref{fig:joint_designs}(b) compares the importance maps predicted using  VISIMP and AGD-F. AGD-F outperforms the VISIMP model and is slightly less accurate than UMSI on GDI ($R$ is the Spearman correlation rank coefficient) per image. AGD-F has mean $R^2$ = 0.732, versus  VISIMP's 0.576 and UMSI's 0.781. The AdamW optimizer with an initial learning rate = 0.9 was used to train this model. The network converged within 15 epochs. See Fig.~\ref{fig:tsnes}(b) for t-SNE plot of the PageEncoder representation on poster images. Based on this t-SNE visualization and elbow of the K-means++ curve (see supplementary), we determined the number of clusters to be 4. We call these clusters: Sparse-Image (with sparse text and images), Text-Heavy (containing mostly text), Image-Text-Complimentary (having a combination of text and images in almost equal proportions) and Singular-Focus (often containing a single big object which attracts attention). In Fig.~\ref{fig:joint_designs}(b), we show the  importance maps predicted by VISIMP \cite{bylinskii2017learning} and our method. In Fig.~\ref{fig:joint_designs}(b), we see that AGD-F produces more accurate importance maps compared to VISIMP, while UMSI slightly better predicts importance than AGD-F. We attribute the better predictive performance of AGD-F (compared to VISIMP) to better saliency prediction of the text (rows 2 and 3) and discriminative region (row 1) components in these images. We did not predict scanpaths on posters and comics as temporal gaze data were unavailable in the corresponding datasets.
\vspace{-1.9mm}
\noindent\subsection{Saliency in Mobile UIs:} We trained AGD on the mobile UI dataset \cite{leiva2020understanding} that contains free-viewing gaze data for 193 mobile UIs. We used 153  images for training and the rest for testing our model. See Fig.~\ref{fig:tsnes}(c) for t-SNE plots of the PageEncoder representations of mobile UIs. Similar to webpages and posters, we determined the number of clusters to be 3. We call these clusters: Box-separated, Text-Stripped and  Image-Overlaid, based on their layouts. In Fig.~\ref{fig:joint_designs}(c), we compare the predicted saliency maps using AGD-F and the SAM-ResNet model \cite{cornia2018predicting} trained on this dataset.  Our model better predicts the saliency map as seen in rows 1 and 3. In row 2, both methods achieve similar performance. We obtained a mean CC score = 0.883, versus SAM-ResNet's 0.831. Recall that AGD-F also outperformed SAM-ResNet on DeepComics. We obtained a mean CC score = 0.621 using the model trained on WebSaliency instead of the mobile UI dataset.  

In Fig.~\ref{fig:scanpaths_mob}, we qualitatively compare the scanpaths obtained on mobile UIs with the same methods used for predicting scanpaths on webpages. The scanpaths predicted by AGD-S again more closely resemble the human scanpaths compared to the other models. In Table.~\ref{tab: quant_scanpath_mob}, we quantitatively evaluate the scanpath prediction performance on this dataset. Our AGD-S method outperforms the compared methods for scanpath prediction on mobile UI images. We observe that the Sequence Scores of the different methods are significantly higher compared to the corresponding scores for webpage scanpath prediction, which indicates that the models predicted scanpaths on mobile UIs more accurately than on webpages. Also, we observed higher inter-observer (IO) consensus among viewers using mobile UIs compared to webpages. 
\vspace{-2.5mm}

\noindent\subsection{Saliency in Natural images:} 
\begin{figure}
\centering
\includegraphics[width=9cm]{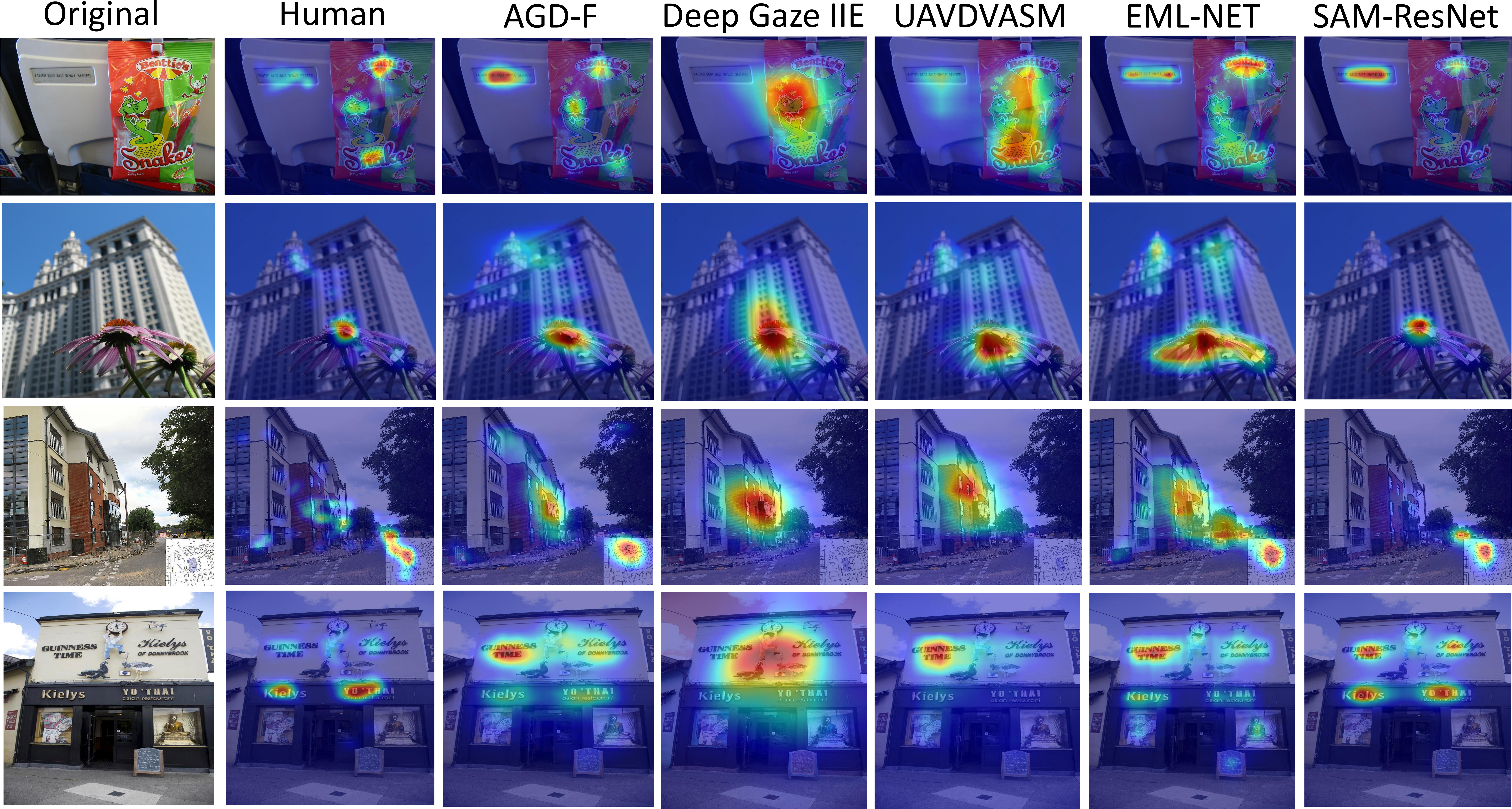}
\caption {Saliency prediction of the AGD-F method and three state-of-the-art natural saliency prediction methods on images from the MIT 1003 dataset.}
\label{fig:natural}
\end{figure}

For evaluating the effectiveness of our AGD-F model for predicting saliency on natural images, we selected the MIT 1003 dataset  \cite{judd2009learning} which contains 1003 natural scene images under different indoor and outdoor settings. In Fig.~\ref{fig:natural} we show the FDMs predicted using the proposed method and four state-of-the-art  natural saliency predictions models Deep Gaze IIE \cite{linardos2021deepgaze}, SAM-ResNet \cite{cornia2018predicting}, UAVDVSM \cite{he2019understanding} and EML-NET \cite{jia2020eml} on some image instances from the MIT 1003 dataset. Our model well predicts saliency on natural images despite being trained only on webpage images. We attribute this success to the fact that several webpage images contain semantically meaningful regions such as faces, people, objects, etc. as part of the website contents, which helps the model well predict the saliency in natural images. In row 4 of Fig.~\ref{fig:natural}, however, we show an image instance where our model fails to accurately predict the saliency of the textual regions on the building. The model inaccurately predicts the top left text on the building to be more salient instead of the more centrally located bottom textual regions, which could be due to the F-bias reading pattern in graphic designs that our model has been trained to capture. This suggests that the F-bias reading pattern in graphic design documents is not always useful for predicting text saliency in natural images. In Table \ref{tab: naturalims}, we quantitatively evaluate the
saliency prediction performance of the compared models on the MIT 1003 dataset. The SAM-ResNet model has the best prediction performance on this dataset.

\begin{table}
\centering
\hfill
\centering
\resizebox{8.6cm}{!}{%
\begin{tabular}{llllll}
\hline\noalign{\smallskip}
 & & & \textbf{MultiMatch} \\
\hline\noalign{\smallskip}
\textbf{Method}    &
\textbf{Sequence Score}    & \textbf{Shape} & \textbf{Direction}
& \textbf{Length} & \textbf{Position} \\
\noalign{\smallskip}
\hline

Itti  & 0.570  &  0.880 &   0.625  &  0.851  & 0.737\\
MASC  & 0.497   &  \textbf{0.924} &  0.675  &  \textbf{0.934}  & 0.706\\
SceneWalker  & 0.632   &  0.893 & 0.674  &  0.883  & 0.790\\
G-Eymol &   0.597   &  0.924  &   0.678 &  0.930  &  0.762\\
AGD-F  &  0.635  &  0.887 &  0.617 &  0.872 & 0.774  \\
AGD-S   &  \textbf{0.664}  &  0.892 & \textbf{0.681}  &  0.881  &  \textbf{0.792}\\
Inter-Observer  & 0.736  &  0.937 &  0.752  &  0.935  & 0.865  \\

\hline
\end{tabular}
}
\caption{Quantitative evaluation of the compared methods for scanpath prediction on the Mobile UI dataset \cite{leiva2020understanding}.}
\label{tab: quant_scanpath_mob}
\vspace{-6mm}
\end{table}

\begin{table}
\centering
\resizebox{8.57cm}{!}{%
\begin{tabular}{llllll}
\hline\noalign{\smallskip}
\textbf{Method}                                & \textbf{NSS $\uparrow$} & \textbf{CC $\uparrow$} & \textbf{KL $\downarrow$}
& \textbf{AUC-J $\uparrow$}  & \textbf{sAUC $\uparrow$}\\ \hline
\noalign{\smallskip}
EML-NET  &  2.518 & 0.677 & 1.433 & 0.377 & 0.656\\ \vspace{0.5mm}
Deep Gaze IIE  &  1.988 & 0.613 & 1.441 &  0.464 & 0.734\\ \vspace{0.5mm}
UAVDVSM  &  2.634 & 0.731 & 1.109 & \textbf{0.475} & 0.776\\ \vspace{0.5mm}
SAM-ResNet  &    \textbf{3.338}                       &      \textbf{0.870}      &        \textbf{0.733}   &  0.422  & \textbf{0.794}\\
AGD-F  &    2.608                       &      0.726       &        1.828    &  0.440  & 0.757\\
\hline
\end{tabular}
}

\caption{Comparison of the proposed method on natural images (MIT 1003 dataset) with natural saliency prediction methods.}
\label{tab: naturalims}
\vspace{-6mm}
\end{table}


\vspace{-1.5mm}
\section{Conclusion}
\label{sec:conclusion}
In this work, we have presented a data-driven model for predicting attention on webpage images and show that our saliency model easily generalizes to other graphic designs such as posters, comics, etc. The model outperforms prior works by combining only the most salient visual components in a webpage to predict saliency. We further show that  utilizing the page layout representation for saliency prediction leads to better performance. Using these layouts we obtain the largest dataset of free-viewing webpage fixations to date. Finally, the component saliency maps are leveraged to predict attention scanpaths on graphic designs.
 
As future work, we plan to develop  a unified model for predicting saliency on different graphic design types \cite{fosco2020predicting}. While we also could have included a straightforward classification sub-network in our model, similar to UMSI \cite{fosco2020predicting}, we rather decided to focus on the primary task of improving attention prediction of different graphic design documents by leveraging component specific FDMs and page layout information. Since our model has been trained on a relatively large dataset of eye-fixations, our saliency model works well for other graphic designs such as comic images (outperforming the popular SAM-ResNet model) and on natural images (as shown in Fig.~\ref{fig:natural} and Table \ref{tab: naturalims} outperforming popular natural saliency models e.g. EML-NET and Deep Gaze IIE) without model re-training. Another future direction could be to leverage the  semantics of a graphic design (e.g. document categories, emotionally engaging faces and texts, etc.) to predict visual attention. This will require collecting more data for capturing such semantic influences. Also, the effect of including audio as another component for predicting attention on audio-visual stimuli (e.g. TV advertisements)  remains to be explored. Audio-visual attention prediction is an emerging topic of research \cite{yao2021deep,min2016fixation,min2020multimodal,min2020study}, since  audio-visual modeling may contribute to better attention prediction on dynamic stimuli. In addition to saliency prediction on graphic designs, the task of quality assessment of screen content images (e.g. compressed natural images and graphic designs) is also an emerging research area  \cite{min2021screen,min2017unified,fu2017screen}. We believe the FDMs and the scanpaths predicted by the proposed model would be useful tools aiding the prediction of screen content quality. While existing literature has focused on predicting saliency in natural videos \cite{wang2018revisiting}, the investigation of attention behavior in dynamic graphic designs remains to be explored. An example for this could be dynamic webpage saliency, where the goal would be to predict the viewing behavior as a webpage loads on the screen. 
 
 \vspace{-0.5mm}


%



 \section*{Acknowledgments}
  This work was supported by the National Science Foundation under grant CNS-1718014.


\ifCLASSOPTIONcaptionsoff
  \newpage
\fi



%

\bibliography{egbib.bib}{}
\bibliographystyle{IEEEtran}

\end{document}